\documentclass[10pt,twocolumn,letterpaper]{article}
\usepackage{authblk}
\usepackage[pagenumbers]{iccv} 
\usepackage{algorithm, algpseudocode}
\usepackage[algo2e]{algorithm2e} 
\usepackage{pifont}
\usepackage[misc]{ifsym}
\usepackage[accsupp]{axessibility}
\usepackage{fancyhdr}
\usepackage{datetime}
\newdateformat{monthyeardate}{%
  \monthname[\THEMONTH], \THEYEAR}

\definecolor{iccvblue}{rgb}{0.21,0.49,0.74}
\usepackage[pagebackref,breaklinks,colorlinks,allcolors=iccvblue]{hyperref}

\def\paperID{1880} 
\def\confName{ICCV}
\def\confYear{2025}

\title{UniOcc: A Unified Benchmark for Occupancy Forecasting and Prediction \\in Autonomous Driving}

\author[ ]{ 
Yuping Wang$^{1,2,4,*}$ 
Xiangyu Huang$^{3, \dagger}$ \quad 
Xiaokang Sun$^{1, \dagger}$ \quad
Mingxuan Yan$^{1}$ \quad
Shuo Xing$^{4}$ \quad
Zhengzhong Tu$^{4}$ \quad
Jiachen Li$^{1, \ddagger}$
}

\affil[ ]{ $^{1}$University of California, Riverside \qquad \qquad \qquad $^{2}$University of Michigan \qquad \qquad}
\affil[ ]{ $^{3}$University of Wisconsin, Madison \qquad \qquad  \qquad \ $^{4}$Texas A\&M University 
}

\usepackage{siunitx}
\usepackage{array, booktabs}
\PassOptionsToPackage{table, svgnames, dvipsnames}{xcolor}
\newcolumntype{P}[1]{>{\centering\arraybackslash}p{#1}}

\newcommand\blfootnote[1]{%
  \begingroup
  \renewcommand\thefootnote{}\footnote{#1}%
  \addtocounter{footnote}{-1}%
  \endgroup
}

\begin{document}

\twocolumn[{%
  \renewcommand\twocolumn[1][]{#1}%
  \maketitle
  \begin{center}
  \vspace{-4mm}
  \includegraphics[width=\linewidth, trim=0 0 0pt 0, clip]{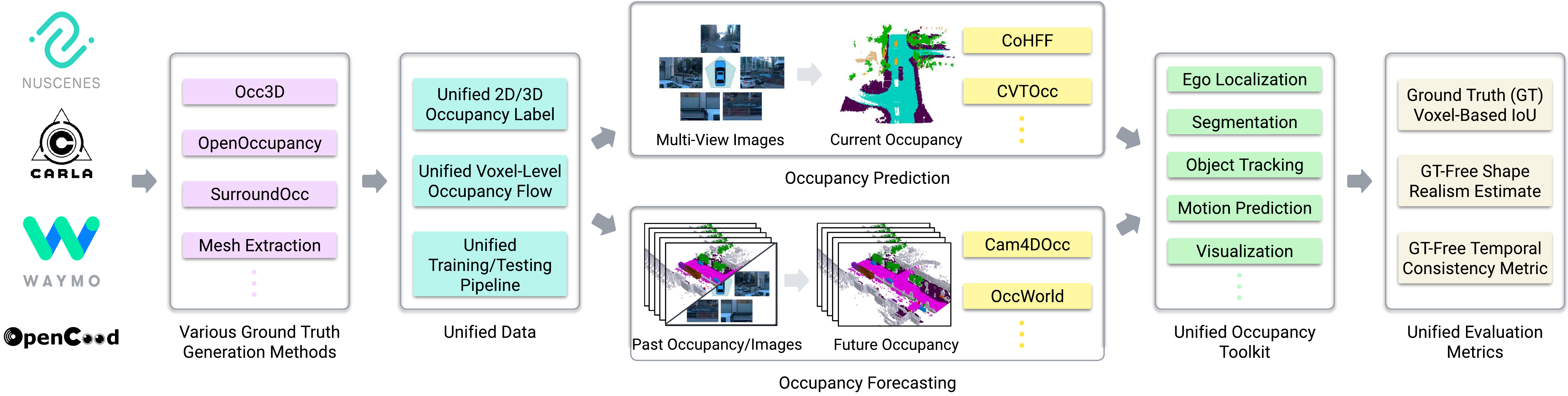}
  \end{center}
  \vspace{-10pt}
  \captionof{figure}{
        Our UniOcc framework incorporates various occupancy label generation methods from multiple data sources, provides the training/testing pipeline \& toolkit for a variety of occupancy tasks, and supports comprehensive evaluation metrics.}
    \vspace{0.3cm}
    \label{fig:teaser}
    \vspace{2.em}
}]

\maketitle
\begin{abstract}
We introduce UniOcc, a comprehensive, unified benchmark and toolkit for occupancy forecasting (\textit{i.e.}, predicting future occupancies based on historical information) and occupancy prediction (\textit{i.e.}, predicting current-frame occupancy from camera images. UniOcc unifies the data from multiple real-world datasets (\textit{i.e.}, nuScenes, Waymo) and high-fidelity driving simulators (\textit{i.e.}, CARLA, OpenCOOD), providing 2D/3D occupancy labels and annotating innovative per-voxel flows. 
Unlike existing studies that rely on suboptimal pseudo labels for evaluation, UniOcc incorporates novel evaluation metrics that do not depend on ground-truth labels, 
enabling robust assessment on additional aspects of occupancy quality. Through extensive experiments on state-of-the-art models, we demonstrate that large-scale, diverse training data and explicit flow information significantly enhance occupancy prediction and forecasting performance. Our data and code are available at \href{https://uniocc.github.io/}{https://uniocc.github.io/}.

\vspace{-10mm}

\end{abstract}

\blfootnote{$^{*}$ypw@umich.edu,  $^{\dagger}$\text{Equal contribution}}
\blfootnote{$^\ddagger$jiachen.li@ucr.edu, Corresponding author}

\section{Introduction}
\vspace{-2mm}
\label{sec:intro}

Occupancy grid map (OGM) has been an effective representation of traffic scenes. It provides a rasterized view of the environment by discretizing the space into a grid of 2D or 3D cells, each indicating the presence or absence of objects such as vehicles, pedestrians, and static obstacles \cite{zhang2024vision}.
Obtaining robust occupancy representations of the dynamic environments is essential for safe motion planning, end-to-end driving systems, and various off-board applications (\textit{e.g.}, data generation for model development). There are two representative tasks in the context of autonomous driving: 
a) Occupancy forecasting \cite{zheng2025occworld, wang2024occsora, bian2025dynamiccity, ma2024cam4docc} aims to predict future occupancies based on historical occupancy/image observations, which enables autonomous systems to anticipate dynamic changes in the environment; b) Occupancy prediction \cite{agro2024uno, ye2024cvtocc} focuses on estimating the current occupancy grid map from raw sensor data, which reconstructs the surrounding scene in a structured and interpretable format.
While recent work has made significant progress, several critical issues have to be addressed.

\textbf{Suboptimal Occupancy Labels and Metrics.} Widely used driving datasets (\textit{e.g.}, nuScenes \cite{nuscenes2019} and Waymo \cite{waymodataset}) lack official occupancy annotations. Existing research thus relies on pseudo ground truth labels derived from heuristics or manual labeling \cite{tian2024occ3d, wei2023surroundocc, wang2023openoccupancy}. These pseudo labels often capture only the reflective surfaces (\textit{e.g.}, car sides hit by LiDAR), failing to represent the true 3D occupancy of the scene. Models trained on these suboptimal labels inevitably produce suboptimal results. Even worse, standard metrics like Intersection-over-Union (IoU) cannot reveal such quality issues because they compare predictions solely against the flawed pseudo labels. To mitigate these pitfalls, we propose novel evaluation metrics that do not rely on pseudo ground truth labels. These metrics provide additional aspects for occupancy quality evaluation.

\textbf{Domain Constraints and Fragmented Data.} Existing occupancy forecasting and prediction methods are mostly restricted to a single dataset. For example, models trained on the nuScenes dataset \cite{nuscenes2019} often are not directly applicable to the Waymo dataset \cite{waymodataset} due to the differences in sensor configurations, data formats, sampling rates, and annotation types. Furthermore, each dataset typically requires its own dedicated tools and data loaders. Inspired by efforts in unified trajectory prediction (\textit{e.g.}, UniTraj \cite{feng2025unitraj}), we introduce a unified occupancy dataset and framework that standardizes these discrepancies, which enables cross-dataset training with a single command. Our framework also leverages CARLA simulations to provide virtually unlimited, diverse training data. Furthermore, our unification enables the cross-domain evaluation of occupancy methods and allows us to analyze their out-of-distribution generalization performance, which is critical for safe autonomous driving.

\textbf{Lack of Per-Voxel Flows.} Current 3D occupancy labels generally lack motion flow information within each voxel, which limits the ability of models to exploit dynamic scene cues. While flows may not be critical in camera-to-occupancy prediction, they are crucial for occupancy forecasting tasks that must capture object and agent movement over time. By including forward and reverse flows for each voxel, our unified dataset facilitates more robust forecasting and simplifies downstream tasks such as object tracking. Furthermore, to our knowledge, we are the first to use per-voxel flows for 3D occupancy forecasting.

\textbf{Lack of Support for Cooperative Occupancy Forecasting.} Cooperative driving is a growing area with research in cooperative perception and prediction \cite{li2024comamba, xu2022cobevt, wang2024cmp, song2024collaborative}, but there has been no dataset available for cooperative occupancy forecasting. Building on OpenCood \cite{xu2022opv2v}, our framework and dataset extend to multi-agent scenarios, serving as the first benchmark to support cooperative occupancy forecasting.

To address these issues, we present UniOcc, a comprehensive, open-source benchmark unifying 2D/3D occupancy labels, per-voxel flow annotations, and multi-agent support across multiple real-world and synthetic datasets. We hope that UniOcc will catalyze occupancy-centric research, streamlining development, benchmarking, and fostering innovations in autonomous driving. The summary of our contributions is as follows:

\begin{itemize}
    \item We introduce UniOcc, the first-of-its-kind unified 2D/3D occupancy forecasting and prediction benchmark with flow information for both conventional and cooperative driving by unifying real data from nuScenes and Waymo and synthetic data from CARLA and OpenCOOD.
    \item We develop a user-friendly platform for current-frame occupancy prediction and multi-frame occupancy forecasting, which enables easy setup, cross-dataset augmentation, and comprehensive occupancy evaluation with or without reference to ground-truth labels.
    \item We provide the Python toolkit for occupancy grid processing: localization, detection, tracking, alignment, and visualization. (See supplementary and code for details.)
    \item  We validate our dataset-agnostic training/testing pipeline and the proposed evaluation metrics on state-of-the-art occupancy forecasting/prediction models. Our experiments show that (1) incorporating voxel-level flow yields performance gains in occupancy forecasting and (2) existing methods face challenges in cross-domain generalization, highlighting avenues for future research.
\end{itemize}

\section{Related Work}
\label{sec:related_works}

\subsection{Occupancy Datasets}
The nuScenes~\cite{nuscenes2019} and Waymo~\cite{waymodataset} datasets are widely used autonomous driving datasets collected from real-world driving, which provide raw sensor data (\textit{i.e.}, camera and LiDAR) with 3D annotations. 
Nevertheless, they do not provide 3D occupancy labels. 
As a result, existing studies often rely on automatic label generation methods introduced in Occ3D \cite{tian2024occ3d}, SurroundOcc \cite{wei2023surroundocc}, or OpenOccupancy \cite{wang2023openoccupancy}. 
On the other hand, CarlaSC \cite{wilson2022motionsc} and CoHFF \cite{song2024cohff} provide synthetic datasets collected with the CARLA simulator \cite{dosovitskiy2017carla}, where the ground truth occupancy can be easily obtained. A detailed comparison between existing datasets and UniOcc (ours) is shown in Table \ref{table:datasets}.

\subsection{Occupancy Prediction}

Recent studies in 3D perception have explored using only camera inputs to produce dense 2D or 3D occupancy grid maps with semantic labels. Early methods often employ a single frame of monocular or multi-camera images to estimate the 2D occupancy \cite{li2023bevdepth,huang2021bevdet,li2022bevformer, toyungyernsub2022dynamics} and 3D occupancy \cite{huang2023tpvformer, wei2023surroundocc, tian2024occ3d, tong2023scene, song2024collaborative, wang2023openoccupancy, ma2024cotr, zhao2024lowrankocc} at the current frame. 
Despite these advances, single-frame methods are limited by their ability to estimate depth. Recently, researchers have turned to historical camera frames for more robust geometric cues, allowing better handling of occlusion and complex scene dynamics. CVT-Occ~\cite{ye2024cvtocc}, for example, refines current-frame occupancy with a \emph{temporal cost volume} constructed from past images, thereby leveraging multi-view images across time for improved depth estimation. In the cooperative autonomous driving domain, CoHFF~\cite{song2024cohff} explores cooperative prediction from a multi-connected vehicle (CAV) setting by having each CAV share its perception information.

\subsection{Occupancy Forecasting} 

Beyond static occupancy reconstruction, a growing line of work tackles \emph{temporal} occupancy prediction, inferring how 3D grids evolve over time. Several recent methods predict future occupancy either from historical occupancy grids~\cite{zheng2025occworld, lange2024self, wang2024occsora,gu2024dome,bian2025dynamiccity,wei2024occllama, ma2024cam4docc, toyungyernsub2024predicting}, often conditioned on ego trajectories or high-level navigation intents. By capturing future scene states, these models facilitate proactive planning and safer driving in dynamic environments.

\subsection{Occupancy Flow}

\begin{figure}[!]
\includegraphics[width=\linewidth]{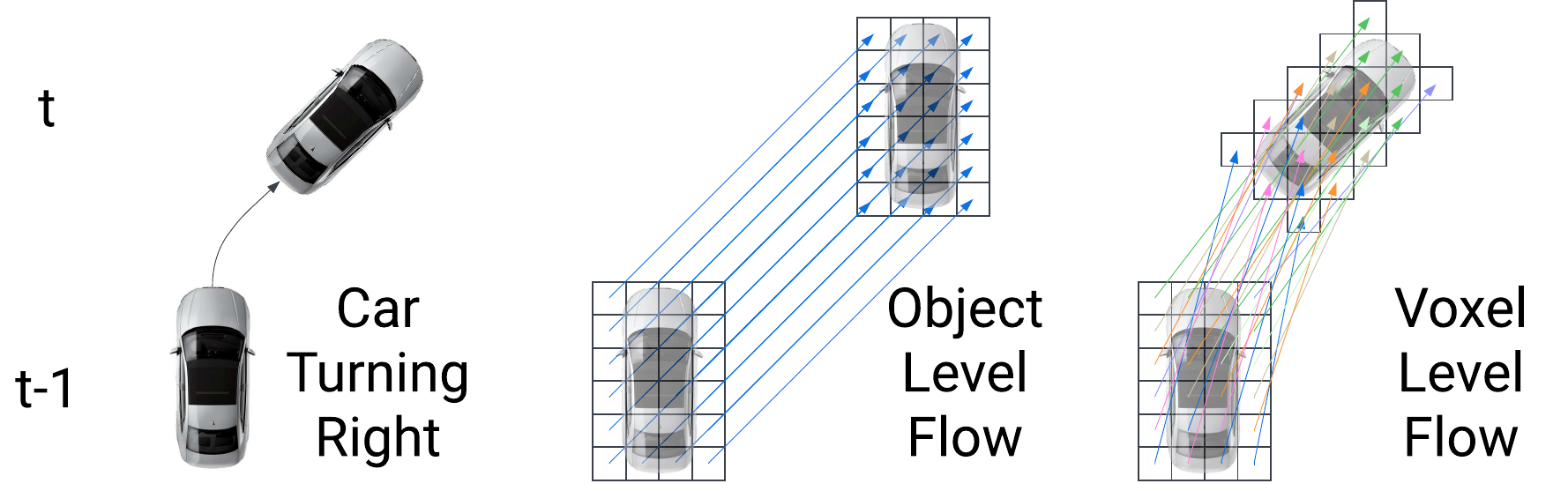}
\centering
\caption{Comparison of object-level flow and voxel-level flow.}
\vspace{-0.2cm}
\label{fig:object_vs_voxel_flow}
\centering
\end{figure}

Early works on flow-driven occupancy forecasting primarily focused on 2D grids, using object bounding boxes and map information to predict future occupancy and flow~\cite{liu2022strajnetflow, liu2023multiflowfield, mahjourian2022occupancyflowfields}. While LetOccFlow~\cite{letoccflow} extends flow to 3D, it only considers horizontal directions and thus cannot capture rich object rotations. CarlaSC~\cite{wilson2022motionsc} provides per-object flow by assigning each voxel the object’s velocity, but this approach similarly neglects rotational motion. In contrast, our method annotates each voxel with its \emph{unique} 3D displacement for the next time step, thereby preserving full rotation and translation. We show their difference in Figure \ref{fig:object_vs_voxel_flow}.

\subsection{Occupancy vs. Trajectory Representations}

Trajectory prediction has been widely studied in autonomous driving and robotics~\cite{ma2021multi, wei2021autonomous, choi2021shared, li2023game, girase2021loki, huang2022survey, lange2024scene, li2020evolvegraph, xie2023cognition, dax2023disentangled, rudenko2020human, ma2022multi, salzmann2020trajectron++, cao2021spectral, wang2023equivariant, xu2024matrix, zhou2022grouptron, yao2025towards, wang2025deployable, zhao2025trajevo}, which aims to predict the future trajectory of dynamic agents (e.g., vehicles, pedestrians, cyclists). However, the trajectory-based methods rely on precise location and are difficult to represent obstacles with arbitrary shapes or geometries \cite{ma2021continual, sun2022interaction, li2023pedestrian, li2024adaptive, li2024multi}. In contrast, occupancy representations model the probabilistic spatial extent of agents over time, naturally capturing multi-modality, uncertainty, and agent-environment interactions~\cite{wang2025deployable}. This motivates the need for a unified benchmark that evaluates forecasting in this richer, more expressive representation space.

Most prior approaches adopt a \textit{forward flow} convention (a vector pointing from the current voxel to its next location). Liu \etal~\cite{liu2023multiflowfield} introduce an alternative \textit{reverse flow} that points backward in time to simplify multi-future training. To accommodate both conventions, we provide both forward and reverse flows in our unified dataset, enabling versatile modeling of complex, fully 3D motion dynamics. 

\begin{table*}[t]
\centering
\footnotesize
  \setlength{\tabcolsep}{1pt}
\caption{Comparison of popular occupancy datasets. \textbf{Length} is the total time this dataset covers. \textbf{Scenarios} are the number of scenarios, usually a proxy for data diversity. \textbf{Voxel Range} is the range of the occupancy grid. \textbf{Resolution} is the per-voxel size. \textbf{Flow} is whether this dataset provides occupancy flow. \textbf{Obj Categories} are the number of category labels provided in the dataset.}
\vspace{-0.3cm}
\begin{tabular}{lcccccccc}
\toprule
\textbf{Dataset} & \textbf{Data Source} & \textbf{Length} & \textbf{Scenarios} & \textbf{Sampling Rate} & \textbf{Voxel Range (\SI{}{m})} & \textbf{Resolution (\SI{}{m})} & \textbf{Flow} & \textbf{Obj Categories}\\ 
\midrule
Occ3D nuScenes~\cite{tian2024occ3d} & nuScenes &\SI{9.5}{hrs} & 1110 & \SI{2}{Hz} & $[\pm40, \pm40, -1\sim5.4]$ & 0.2 / 0.4 & - & 17\\
Occ3D Waymo~\cite{tian2024occ3d} & Waymo & \SI{4.0}{hrs} & 998 & \SI{10}{Hz} & $[\pm40, \pm40, -1\sim5.4]$  & 0.2 / 0.4 & - & 15\\
SurroundOcc~\cite{wei2023surroundocc} & nuScenes & \SI{9.5}{hrs} &1110 & \SI{2}{Hz} &$[\pm40, \pm40, -1\sim5.4]$ & 0.5 & - & 17 \\
OpenOccupancy~\cite{wang2023openoccupancy} & nuScenes & \SI{9.5}{hrs} & 1110 & \SI{2}{Hz}& $[\pm51.2, \pm51.2, -5\sim3]$ & 0.1 & - & 17 \\
CoHFF~\cite{song2024cohff} & OpenCOOD &  \SI{0.69}{hrs} & 44 & \SI{10}{Hz} &$[\pm51.2, \pm51.2, -5\sim3]$ & 1.0 & - & 10 \\
\midrule
UniOcc (Ours) & \begin{tabular}{@{}c@{}}nuScenes, Waymo\\ CARLA, OpenCOOD\end{tabular} & \SI{14.2}{hrs} & 2152 & \SI{2}{Hz} / \SI{10}{Hz} &$[\pm40, \pm40, -1\sim5.4]$ & 0.2 / 0.4 & Voxel Level & 10, 15, 17 \\
\bottomrule
\end{tabular}
\label{table:datasets}
\end{table*}

\section{UniOcc Framework}
\label{methods}

\subsection{Unified Data Format and Features}

Our benchmark supports a wide range of occupancy-centric tasks, including occupancy forecasting, single-frame occupancy prediction, and flow estimation. Our framework defines the following task-agnostic data formats:

\noindent\textbf{Semantic Occupancy Label.}\label{occ_label}
We represent the scene as a 3D voxel grid $G \in \{0, \ldots, C\}^{L \times W \times H}$, where $C$ denotes the number of classes (see Table~\ref{table:label_comparison} in supplementary materials), and $L$, $W$, $H$ are the grid’s dimensions along the ego vehicle’s heading, lateral, and vertical axes, respectively. This grid is centered on the ego vehicle, with the $+x$-axis aligned to the direction of travel, $+y$-axis to the left, and $+z$-axis upward. For certain 2D tasks (\textit{e.g.}, motion planning), we collapse the height dimension via a priority scheme (\textit{e.g.}, \textit{Pedestrian} $>$ \textit{Car} $>$ \textit{Road}), such that each vertical pillar adopts the label of its highest-priority voxel. This approach prevents occlusion of essential object classes (like pedestrians) by lower-priority labels in the same grid column, ensuring meaningful representation for downstream tasks.

\noindent\textbf{Camera Images.}
We store raw RGB images in a 4D tensor $
I \in \{0,\dots,255\}^{K_{\text{cam}} \times \text{Img}_x \times \text{Img}_y \times 3}$, where $K_{\text{cam}}$ denotes the number of onboard cameras and each image has resolution $\text{Img}_x \times \text{Img}_y$.

\noindent\textbf{Camera Field-of-View (FOV) Mask.}
A binary 3D tensor $U \in \{0,1\}^{L \times W \times H}$
indicates which voxels lie within each camera’s observable frustum ($U=1$ for visible voxels and $U=0$ otherwise). This mask is crucial for camera-based occupancy methods that require explicit delineation of occluded regions or unobserved space.

\noindent\textbf{Camera Intrinsics and Extrinsics.}
We represent camera intrinsics as $
\text{Int} \in \mathbb{R}^{K_{\text{cam}} \times 3}$, 
while extrinsic transformations (from each camera to the ego frame) are given by $
\text{Ext} \in \text{SE}(3)^{K_{\text{cam}}}$, where $\text{SE}(3)$ denotes the group of 3D homogeneous transformation. These parameters unify the projection from 3D ego coordinates onto 2D image planes.

\noindent\textbf{Ego-to-World Transformation.}
A homogeneous transformation matrix $T_e^w \in \text{SE}(3)$
denotes the pose of the ego vehicle in a global world frame, enabling precise alignment of data from multiple sensors and coordinate systems.

\begin{figure*}[!tbp]
\includegraphics[width=0.9\textwidth]{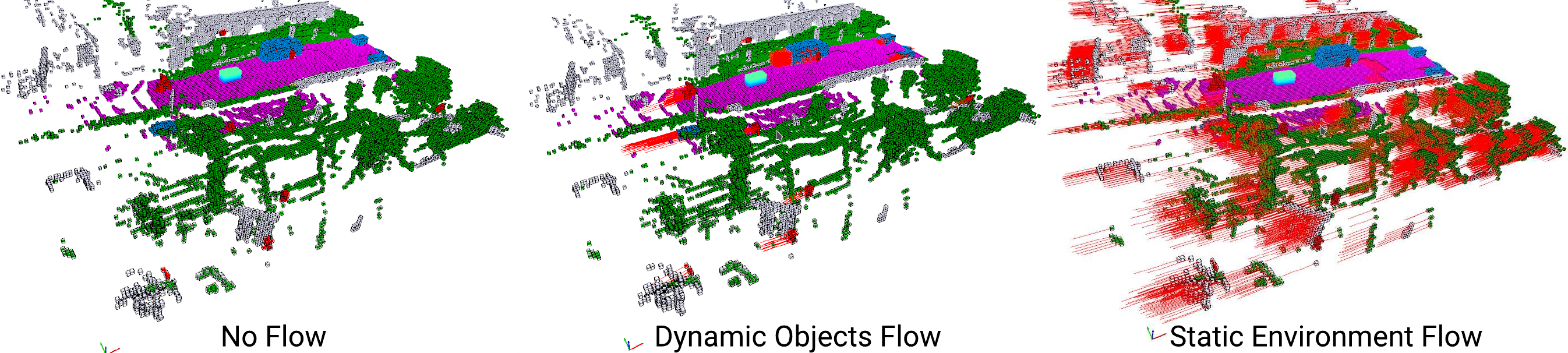}
\centering
\caption{Visualization of the per-voxel forward flows.}
\label{fig:flow_demo}
\centering
\end{figure*}

\noindent\textbf{Forward Occupancy Flow.}\label{sec:flow} We define a 4D tensor $F \in \mathbb{R}^{L \times W \times H \times 3}$ that records per-voxel forward-motion vectors. Unlike prior methods~\cite{wilson2022motionsc} that assign a single velocity to all voxels of an object (thus missing object rotation), our approach computes individual voxel flows capturing both translation and rotation. We separately compute the flow for dynamic foreground objects (\textit{e.g.}, \textit{Car}, \textit{Pedestrian}) and static background environment (\textit{e.g.}, \textit{Road}, \textit{Vegetation}) and merge dynamic and static flows into $F$. As illustrated in Figure~\ref{fig:flow_demo}, this voxel-level flow captures full 3D motion, including rotation. The details of our computation algorithm can be found in the supplementary materials and code. 

\noindent\textbf{Backward Occupancy Flow.}
Analogous to the forward flow, we define a 4D tensor $B \in \mathbb{R}^{L \times W \times H \times 3}$ to capture \emph{backward} motion vectors. Instead of computing each voxel’s displacement from $t$ to $t+1$, we evaluate the motion from $t$ to $t-1$. This backward flow is particularly useful for models that benefit from reverse-time supervision or multi-future training strategies~\cite{liu2023multiflowfield}.

\noindent\textbf{Object Annotations.}
We also provide object-level annotations as a list of dictionaries, each containing:
    \textbf{\ding{182} Agent-to-Ego Transformation.} A transformation matrix $T_a^e \in \text{SE}(3)$ that maps the agent’s local coordinate system into the ego frame. This captures both the agent’s position and orientation relative to the ego vehicle.
    \textbf{\ding{183} Size.} A 3D vector $d \in \mathbb{R}^3$ describing the bounding box dimensions of the agent (\textit{length}, \textit{width}, \textit{height}).
    \textbf{\ding{184} Category.} The object’s semantic class label, following the definitions in Table~\ref{table:label_comparison}.

\subsection{Task Categories}
\vspace{-1mm}

Our unified inputs enable a broad range of occupancy-centric tasks, spanning both static prediction and dynamic forecasting. By employing a unified representation across multiple domains, we simplify cross-dataset training and allow fair comparisons of methods that tackle different sub-problems. Below, we outline three representative tasks:

\noindent\textbf{Occupancy Prediction.}
Here, the model consumes the past $W_{\text{obs}}$ camera frames $\{I^{t-W_{\text{obs}},\dots,t}\}$, together with their FOV masks $\{U^{t-W_{\text{obs}},\dots,t}\}$ and camera parameters (\textit{intrinsics} $\text{Int}$, \textit{extrinsics} $\text{Ext}$). The output is the current 3D occupancy grid $G^t$, which captures the scene at time $t$. 

\noindent\textbf{Occupancy Forecasting with Optional Flow.}
\label{forecast_definition} In the forecasting setting, the input is the historical data over $W_{\text{obs}}$ frames—either voxel grids $\{G^{t-W_{\text{obs}},\dots,t}\}$ or camera images $\{I^{t-W_{\text{obs}},\dots,t}\}$. The model predicts future occupancies $\{G^{t,\dots,t+W_{\text{fut}}}\}$, optionally conditioned on fine-grained ego trajectories $T_{e}^{w,t:\,t+W_{\text{fut}}}$ or high-level driving intentions (\textit{e.g.}, \textit{Turn Right}). For certain use cases, forecasting methods may also produce the future flow $F^{t:\,t+W_{\text{fut}}}$ or future ego movement $T_{e}^{w,t:\,t+W_{\text{fut}}}$. As discussed in Section~\ref{flow_exp}, this joint occupancy-and-flow forecasting scheme can help capture complex motion patterns over time.

\noindent\textbf{Cooperative Occupancy Prediction and Forecasting with Optional Flow.}
Under cooperative settings, multiple connected vehicles (CAVs) collaborate by sharing either image or occupancy data. From the ego vehicle’s perspective, it receives the shared historical observations $\{I_{\text{CAV}}^{t-W_{\text{obs}},\dots,t}\}$ or $\{G_{\text{CAV}}^{t-W_{\text{obs}},\dots,t}\}$ alongside transformations mapping CAV frames to the ego frame. The output remains the same (\textit{i.e.}, single-ego occupancy or forecast), but the increased viewpoint coverage can mitigate occlusions and improve overall scene understanding.

\subsection{Unified Datasets}
\vspace{-1mm}

We build our unified dataset from the following sources:

\begin{itemize}
    \item \textbf{nuScenes}~\cite{nuscenes2019} \textbf{and Waymo}~\cite{waymodataset}. 
    Both datasets provide camera images, LiDAR sweeps, and object-level annotations. As neither directly includes 3D occupancy labels, we synthesize occupancy ground truth via three pseudo-labeling pipelines: Occ3D~\cite{tian2024occ3d}, OpenOccupancy~\cite{wang2023openoccupancy}, and SurroundOcc~\cite{wei2023surroundocc}. This multi-tool approach increases robustness and diversity in labeled outputs.

    \item \textbf{CARLA}~\cite{dosovitskiy2017carla}. 
    We use CARLA’s simulation engine to generate an unlimited variety of virtual driving scenes, from which we can extract “perfect” 3D occupancy labels (meshes, object states, etc.). These realistic yet controllable scenarios are publicly released, enabling straightforward large-scale training. Our framework offers the option to generate an arbitrary length of data.

    \item \textbf{OpenCOOD}~\cite{xu2022opv2v}. 
    Built on CARLA, OpenCOOD offers multi-vehicle cooperation scenarios. We extend its data-generation scripts to export 3D occupancy from simulated meshes, thus expanding our dataset with collaborative driving examples.
\end{itemize}

\vspace{-0.2cm}
\subsection{Unified Occupancy Processing Toolkit}
\vspace{-1mm}
\label{sec:unified_toolkit}

\begin{figure}[!]
    \centering
    \includegraphics[width=0.46\textwidth]{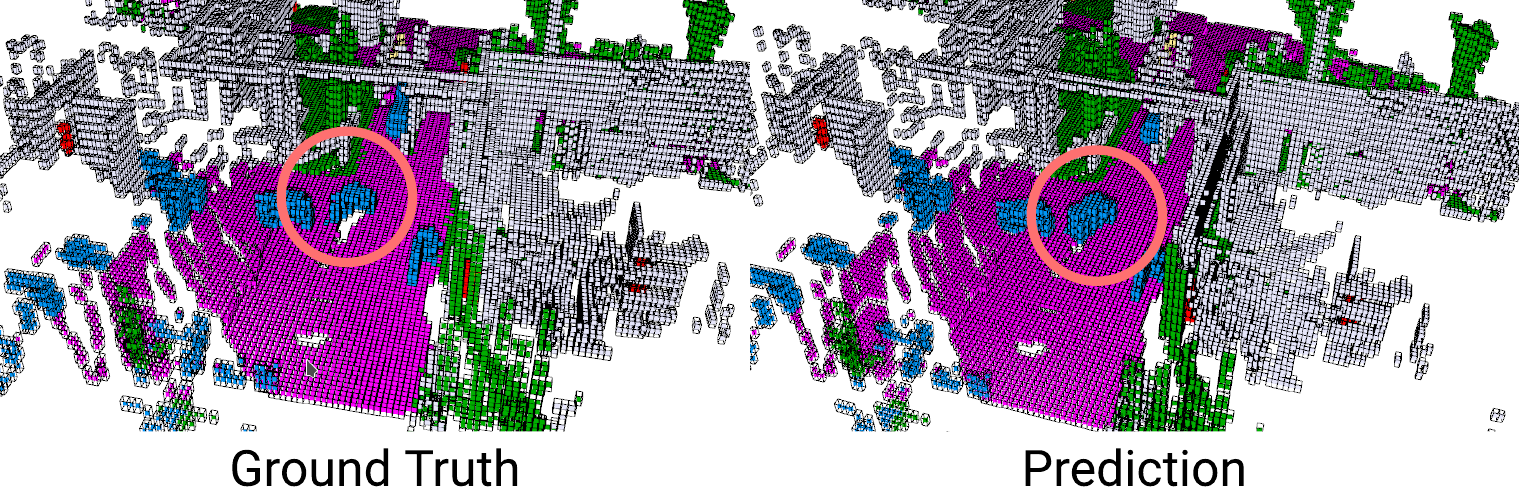}
    \caption{Illustration of imperfect ground-truth labels. Left: partial car shape from Occ3D~\cite{tian2024occ3d}. Right: a more complete shape predicted by OccWorld~\cite{zheng2025occworld}. Standard IoU may penalize the model for producing a fuller shape, despite it being more realistic.}
    \label{fig:gt_flaw}
\end{figure}

Most occupancy-based approaches focus solely on generating an occupancy grid but provide limited tools for downstream processing or motion analysis. To address this gap, our framework includes a toolkit for object segmentation and tracking directly within the voxel space, enabling more advanced tasks such as shape analysis or motion planning. (Some details, \textit{e.g.}, localization, tracking, are deferred to the supplementary and the code).

\subsubsection{Object Identification}
\label{sec:object_identification}
\begin{figure*}[t]
\centering
\includegraphics[width=0.95\textwidth]{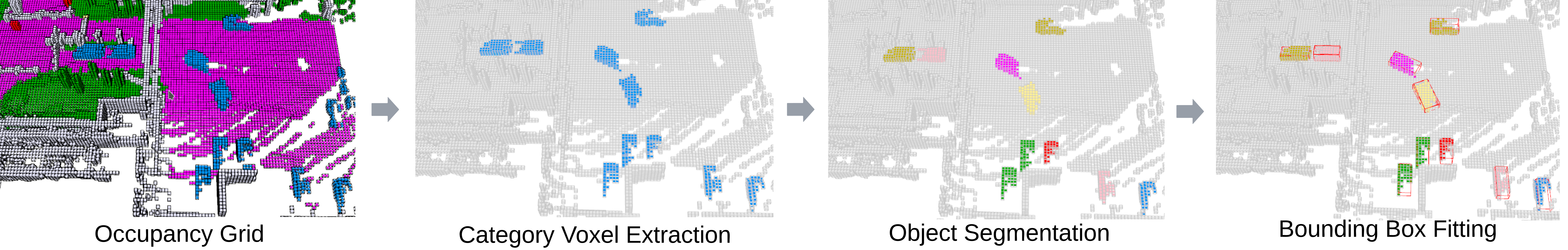}
\caption{An example showing our pipeline for voxel extraction, connected-component segmentation, and bounding-box fitting.}
\label{fig:seg_process}
\end{figure*}

Given an occupancy grid \(G \in \{0,\dots,C\}^{L \times W \times H}\), we identify and segment relevant objects in the following steps, which are shown in Figure~\ref{fig:seg_process}.
\begin{enumerate}[leftmargin=*]
    \item \textbf{Object Segmentation.}
    We extract voxels by category (\textit{e.g.}, \textit{Car}, \textit{Pedestrian}), then run 6-connected component labeling (CCL) implemented via Breadth-First Search:
    \begin{equation}
        L = \mathrm{CCL}(G), \quad t \in \{0,1,\dots,T\},
    \end{equation}
    where \(L \in \{1,\dots,N\}^{L \times W \times H}\) assigns each connected component a unique object ID, and \(N\) is the total number of objects.

    \item \textbf{Voxel Extraction.}
    For each object ID \(n\), we gather its voxel coordinates $V_n$:
    \begin{equation}         
    \label{eq:argwhere}
        V_n=\{<x,y,z>|L(x,y,z)=n\}
    \end{equation}

    \item \textbf{Horizontal Axis Bounding Box.}
    Voxel predictions can be partial (see Fig.~\ref{fig:gt_flaw}), making direct bounding-box measurement (length, width, height) unreliable. We therefore fit a bounding rectangle in the horizontal plane using a rotating-calipers method~\cite{toussaint1983solving}, which is \(O(n^2)\) in the number of object voxels, under the assumption that each object moves parallel to the ground. This yields a 2D minimum bounding rectangle, from which we recover heading and planar extents.

    \item \textbf{Dimension Extraction.}
    We take the rectangle’s length and width as the object’s planar dimensions, then compute height from the vertical extent of the voxels. All dimensions are scaled by the voxel resolution \(\epsilon\) to convert to metric units.
\end{enumerate}

\subsubsection{Object Tracking}
\label{sec:object_tracking}
Leveraging the forward occupancy flow introduced in Section \ref{sec:flow} predicted for each voxel, we also provide a simple occupancy-based object tracking algorithm:
\begin{enumerate}[leftmargin=*]
    \item \textbf{Object Voxel Extraction.}
    \label{extraction}
    For each identified object in the occupancy grid at frame \(t\), we retrieve its voxel coordinates \(V_n^t\) (Eq.~\ref{eq:argwhere}) and corresponding flow vectors \(F_n^t \in \mathbb{R}^{n \times 3}\).

    \item \textbf{Step Prediction.}
    We estimate the next-frame voxel positions $\widetilde{V_n^{t+1}}$ by adding the flow:
    \begin{equation}         
    \widetilde{V_n^{t+1}} = V_n^t + F_n^t.
    \end{equation}

    \item \textbf{Centroid Extraction.}
    Let \(\widetilde{c_n^{t+1}}\) be the centroid of the predicted voxel set \(\widetilde{V_n^{t+1}}\). We also compute the true object voxels at frame \(t+1\), \(V_n^{t+1}\), and its centroid \(c_n^{t+1}\):
    \begin{equation}         
    \label{centroid_computation}
        \widetilde{c_n^{t+1}} 
        = \frac{1}{\left|\widetilde{V_n^{t+1}}\right|}\sum_{(i,j,k)\in \widetilde{V_n^{t+1}}} <i,j,k>, 
     \end{equation}
     \begin{equation}     
        c_n^{t+1} 
        = \frac{1}{\left|V_n^{t+1}\right|}\sum_{(i,j,k)\in V_n^{t+1}} <i,j,k>.
    \end{equation}

    \item \textbf{Bipartite Association.}
    We match predicted centroids \(\{\widetilde{c_p^t}\}\) with observed centroids \(\{c_q^{t+1}\}\) using the Hungarian algorithm to minimize pairwise distances:
    \begin{equation}
        P^* 
        = \arg\!\min_P 
        \sum_{p,q} \|\widetilde{c_p^t} - c_q^{t+1}\|_2 \, P_{pq}, 
        \ P \in \{0,1\}^{n_s^t \times n_s^{t+1}}.
    \end{equation}
    where $P^*$ is the matching matrix and $n_s^t$, $n_s^{t+1}$ represents the number of objects in the consecutive frames. We assign the same ID for matched objects between frames and assign new IDs for newly appeared objects. We discard objects where the matched distance is greater than a threshold of $\gamma$, which we empirically choose as \SI{0.5}{m} for cars and cyclists, \SI{0.2}{m} for pedestrians.

\end{enumerate}

\noindent The above process yields cross-frame associations that unify object identities over time, enabling motion interpretation and analysis directly in the voxel space. 

\subsubsection{Object Alignment}
\label{pca}
Finally, we align the voxel sets of tracked objects for shape analysis or consistency checks:
\begin{enumerate}[leftmargin=*]
    \item \textbf{Translation Alignment.}
    We translate each object’s voxel coordinates to center them at the origin as $\bar{V_n^t}$:
    \begin{equation}
    \bar{V_n^t} 
    = V_n^t 
    - \frac{1}{|V_n^t|}\sum_{\mathbf{v} \in V_n^t}\mathbf{v}.
    \end{equation}
    \item \textbf{Rotation Alignment.}\label{sec:pca}
    We apply Principal Component Analysis (PCA) to each frame’s voxel set to resolve a canonical orientation. For consistency, we adjust the sign of the new principal axes to align with the previous frame’s orientation (see supplementary materials). The final rotated voxel coordinates are denoted as \(\hat{V_n^t}\).
\end{enumerate}

\noindent With these steps, we facilitate object-centric analyses (\textit{e.g.}, measuring shape changes or rotation consistency) entirely in the occupancy grid domain without the need for a reference ground truth label or annotation.

\subsection{Unified Evaluation Metrics}
\label{sec:unified_eval}

Our benchmark includes multiple metrics for assessing the quality of generated or predicted occupancy grids. Section~\ref{voxel_metrics} describes the widely adopted \emph{voxel-based metrics}, while Section~\ref{gt_free_eval} proposes \emph{ground-truth-free} methods that address two major issues: the imperfect nature of real-world labels (Fig.~\ref{fig:gt_flaw}) and the inherent multi-modality of some forecasting tasks (where only a single future is recorded but we expect model to produce multiple futures).

\subsubsection{Voxel-Based Evaluation}
\label{voxel_metrics}
Following prior occupancy prediction~\cite{ye2024cvtocc} and forecasting~\cite{wei2024occllama, zheng2025occworld, wang2024occsora, bian2025dynamiccity} studies, we employ two standard metrics: \emph{geometric IoU} (or simply \(\text{IoU}_{\text{geo}}\)) and \emph{mIoU} (mean intersection over union across semantic classes). Concretely, for a predicted occupancy grid \( G_{\text{pred}} \) and ground-truth grid \( G_{\text{gt}} \),
\small
\begin{equation}
\label{eq:iou_def}
    \text{IoU}_{\text{geo}} 
    = \frac{\lvert G_{\text{pred}} \,\cap\, G_{\text{gt}} \rvert}{\lvert G_{\text{pred}} \,\cup\, G_{\text{gt}} \rvert},
\end{equation}
\normalsize
where \(\lvert G_{\text{pred}} \,\cap\, G_{\text{gt}} \rvert\) is the number of voxels occupied in both the prediction and ground truth, and \(\lvert G_{\text{pred}} \,\cup\, G_{\text{gt}} \rvert\) is the total occupied voxels in either grid. For multi-class occupancy (\(C\) total classes), the mIoU is computed via:
\small
\begin{equation}
    \text{mIoU}_{\text{geo}} 
    = \frac{1}{C}\sum_{c=1}^{C} \text{IoU}_{\text{geo},c},
\end{equation}
\normalsize
where \(\text{IoU}_{\text{geo},c}\) is computed from Eq.~\eqref{eq:iou_def} with restricting the voxels to class \(c\).

While these voxel-based metrics are straightforward, they can penalize predictions that exceed the pseudo-ground truth (Fig.~\ref{fig:gt_flaw}). Additionally, tasks like multi-modal forecasting may produce many plausible futures not captured by a single reference label. For these reasons, we propose evaluation strategies that do \emph{not} require perfect ground truth.


\begin{table*}[h!]
\centering
\caption{Occupancy forecasting performance of OccWorld \cite{zheng2025occworld} on nuScenes  with different types of flow. (See Supplementary for Waymo.)}
\vspace{-0.3cm}
\label{table:occworld_flow}
\resizebox{\textwidth}{!}{
\begin{tabular}{P{44mm}|P{17mm}|cccc|cccc|ccc}
\toprule
\textbf{Train and Test Source} & \textbf{Flow Type} & 
\multicolumn{4}{c|}{$\textbf{mIoU}_{\text{geo}}$\(\uparrow\)} &
\multicolumn{4}{c|}{$\textbf{IoU}_{\text{geo}}$\(\uparrow\)} &
\multicolumn{1}{c}{$\textbf{IoU}_{\text{bg}}$\(\uparrow\)} &
\multicolumn{1}{c}{$\textbf{IoU}_{\text{car}}$\(\uparrow\)} &
\multicolumn{1}{c}{$\textbf{P}_{\text{car}}$\(\uparrow\)} \\

& & \SI{0}{s} & \SI{1}{s} & \SI{2}{s} & \SI{3}{s} &
\SI{0}{s} & \SI{1}{s} & \SI{2}{s} & \SI{3}{s} &
\multicolumn{3}{c}{\text{Average over } \SIrange{0}{3}{s}}  \\
\midrule

nuScenes & None &
66.79 & 30.23 & 21.67 & 18.13 &
60.66 & 33.33 & 24.93 & 20.67 &
53.27 & 78.39 & 77.83 \\

nuScenes & Object &
65.41 & 31.09 & 20.97 & 18.40 &
60.72 & 33.28 & 24.57 & 20.55 &
54.15 & 77.36 & 76.25 \\

nuScenes & Voxel &
\textbf{70.64} & \textbf{32.13} & \textbf{22.50} & \textbf{19.06} &
\textbf{62.62} & \textbf{35.93} & \textbf{26.03} & \textbf{21.04} &
\textbf{59.56} & \textbf{81.50} & \textbf{82.57} \\

\bottomrule
\end{tabular}
}
\end{table*}

\begin{table*}[h!]
\centering
\vspace{0.3cm}
\caption{Occupancy forecasting performance of OccWorld \cite{zheng2025occworld} on various train/test data source combinations. }
\vspace{-0.3cm}
\label{table:occworld}
\resizebox{\textwidth}{!}{
\begin{tabular}{c|c|cccc|cccc|ccc}
\toprule
\textbf{Train Sources} & \textbf{Test Source} & 
\multicolumn{4}{c|}{$\textbf{mIoU}_{\text{geo}}$\(\uparrow\)} &
\multicolumn{4}{c|}{$\textbf{IoU}_{\text{geo}}$\(\uparrow\)} &
\multicolumn{1}{c}{$\textbf{IoU}_{\text{bg}}$\(\uparrow\)} &
\multicolumn{1}{c}{$\textbf{IoU}_{\text{car}}$\(\uparrow\)} &
\multicolumn{1}{c}{$\textbf{P}_{\text{car}}$\(\uparrow\)} \\
& & \SI{0}{s} & \SI{1}{s} & \SI{2}{s} & \SI{3}{s} &
\SI{0}{s} & \SI{1}{s} & \SI{2}{s} & \SI{3}{s} &
\multicolumn{3}{c}{\text{Average over } \SIrange{0}{3}{s}}  \\
\midrule

nuScenes & nuScenes &
70.64 & 32.13 & 22.50 & 19.06 &
62.62 & 35.93 & 26.03 & 21.04 &
59.56 & 81.50 & 82.57 \\

Waymo & nuScenes &
63.22 & 23.47 & 18.11 & 15.80 &
60.42 & 27.35 & 20.86 & 17.63 &
49.90 & 79.41 & 72.54 \\

CARLA & nuScenes &
29.93 & 11.94 & 10.85 & 10.54 &
49.32 & 13.51 & 11.24 & 10.82 &
22.39 & 59.64 & 78.99 \\

nuScenes & Waymo &
64.37 & 31.08 & 23.48 & 20.90 &
65.38 & 39.38 & 30.94 & 27.40 &
61.25 & 81.34 & 72.69 \\

Waymo & Waymo &
71.35 & 32.04 & 25.77 & 23.76 &
72.69 & 36.04 & 30.48 & 27.96 &
58.26 & 89.30 & 86.68 \\

CARLA & Waymo &
30.64 & 12.09 & 11.13 & 10.79 &
55.99 & 16.16 & 13.84 & 13.07 &
23.05 & 57.05 & 77.18 \\

nuScenes & CARLA &
79.62 & 49.70 & 49.25 & 48.72 &
68.93 & 17.62 & 16.45 & 15.50 &
86.74 & 97.52 & 71.02 \\

Waymo & CARLA &
80.32 & 48.54 & 48.06 & 47.60 &
71.38 & 15.72 & 14.17 & 12.99 &
86.79 & 91.37 & 80.38 \\

CARLA & CARLA &
79.66 & 48.87 & 47.28 & 46.69 &
69.67 & 20.05 & 15.34 & 12.78 &
24.34 & 59.39 & 80.92 \\

\midrule

nuScenes + Waymo & nuScenes &
71.80 & 32.57 & 22.87 & 20.15 & 
63.12 & 36.24 & 27.93 & 21.48 & 
60.23 & 86.08 & 83.14 \\

nuScenes + Waymo & Waymo &
71.23 & 33.42 & 26.52 & 24.99 & 
73.13 & 36.30 & 31.61 & 28.22 & 
59.23 & 88.72 & 87.01 \\

\midrule

nuScenes + CARLA & nuScenes &
71.47 & 31.70 & 22.69 & 18.11 &
62.94 & 35.83 & 28.29 & 21.03 &
55.07 & 77.87 & 82.97 \\

Waymo + CARLA & nuScenes &
64.99 & 24.88 & 19.65 & 15.55 &
61.33 & 27.50 & 20.38 & 18.94 &
49.20 & 81.13 & 82.50 \\

nuScenes + CARLA & Waymo &
65.71 & 31.48 & 23.84 & 21.25 &
66.95 & 40.37 & 31.47 & 27.50 &
57.34 & 83.80 & 81.98 \\

Waymo + CARLA & Waymo &
71.66 & 37.05 & 29.50 & 26.16 &
72.94 & 42.54 & 35.46 & 31.52 &
56.32 & 86.67 & 81.29 \\

nuScenes + CARLA & CARLA &
84.88 & 49.25 & 48.69 & 47.88 &
74.15 & 17.02 & 15.81 & 14.41 &
86.54 & 98.48 & 81.91 \\

Waymo + CARLA & CARLA &
83.81 & 54.31 & 52.94 & 52.13 &
74.34 & 27.42 & 24.60 & 23.04 &
73.89 & 90.47 & 81.48 \\

\midrule

nuScenes + Waymo + CARLA & nuScenes &
72.53 & 33.98 & 22.76 & 20.18 &
63.32 & 36.31 & 27.83 & 21.89 &
57.51 & 80.51 & 83.01 \\

nuScenes + Waymo + CARLA & Waymo &
74.49 & 34.32 & 28.28 & 24.61 &
73.58 & 43.42 & 32.46 & 27.44 &
62.20 & 87.54 & 80.93 \\

nuScenes + Waymo + CARLA & CARLA &
85.26 & 55.19 & 52.58 & 50.96 &
74.63 & 28.33 & 22.31 & 19.35 &
74.15 & 88.61 & 82.35 \\

\bottomrule
\end{tabular}
}
\end{table*}


\subsubsection{Ground-Truth-Free Evaluation}
\label{gt_free_eval}
Complementary to label-dependent IoU, we propose metrics that assess geometric plausibility without referencing a single ground-truth scene. These metrics are particularly useful for multi-modal generation or cases where ground-truth labels are incomplete.

\noindent\textbf{Key Object Dimension Probability.}
Given a predicted object’s bounding box $<l,w,h>$ for a category \(c\), we evaluate its plausibility by computing a \emph{Gaussian Mixture Model} (GMM) likelihood. Specifically, each category \(c\) has a pretrained GMM, denoted \(\text{GMM}_c\), learned from real or synthetic data in our unified dataset (Appendix~\ref{appendix:gmm}). At inference time, we query \(\text{GMM}_c\) with the object’s dimensions:
\begin{equation}
    P_n = \max_k\, p\bigl(<l,w,h> \,\bigm|\text{GMM}^k_c\bigr),
\end{equation}
This probability \(P_n\) gives a heuristic for whether the object has realistic dimensions for its reported category. We use an empirical value $\rho=0.5$ as the threshold to determine if the shape is likely real or not. 

\noindent\textbf{Temporal Foreground Object Shape Consistency.}
\label{shape_consistency}
For dynamic objects forecasted across multiple frames, we measure shape consistency by aligning each object’s voxels over time (see Section~\ref{pca}) and computing a voxel-wise intersection over union:
\small
\begin{equation}
    \text{IoU}_\text{object} 
    = \frac{\lvert \hat{V}^{t}\,\cap\, \hat{V}^{t+1}\rvert}{\lvert \hat{V}^{t}\,\cup\, \hat{V}^{t+1}\rvert}.
\end{equation}
\normalsize
A high \(\text{IoU}_\text{object}\) implies stable shape geometry from frame \(t\) to \(t+1\). We then average these IoUs within each category to assess overall consistency across time.

\noindent\textbf{Temporal Background Environment Consistency.}
For static background regions, we expect persistent occupancy between consecutive frames within the overlapping field of view. Let \(V_e^t\) be the environment voxels at time \(t\), and \(\widetilde{V}_{e}^{t+1}\) be their projected coordinates at \(t+1\) (using known ego-motion, see Section~\ref{forecast_definition}). We discard out-of-bound voxels and compute the binary IoU of the overlap:
\small
\begin{equation}
    \text{IoU}_{\text{bg}} 
    = \frac{\lvert \widetilde{V}_{e}^{t+1}\,\cap\, V_{e}^{t+1}\rvert}{\lvert \widetilde{V}_{e}^{t+1}\,\cup\, V_{e}^{t+1}\rvert}.
\end{equation}
\normalsize
Higher \(\text{IoU}_{\text{bg}}\) indicates a consistent static background across frames, even without a perfect ground truth label.

Overall, these ground-truth-free metrics complement standard IoU by providing deeper insights into scene realism and temporal coherence, especially valuable for generative or multi-modal occupancy tasks.

\section{Experiments}
\label{experiments}

\subsection{Experimental Settings}

In all experiments, we use a voxel size of $[200, 200, 16]$ and grid resolution of \SI{0.4}{m}. 
For the occupancy forecasting task, the model takes in the \SI{3}{s} historical occupancy and forecasts the future \SI{3}{s}. For the occupancy prediction task, the models take in the \SI{3}{s} historical camera images and predict the occupancy at the current frame.
For nuScenes and Waymo data sources, we leverage the pseudo occupancy labels from Occ3D \cite{tian2024occ3d}. 
For the CARLA data source, we generate simulated driving data in 16 diverse scenarios.

\subsection{Occupancy Forecasting with Flows}
\label{flow_exp}
To investigate the impact of explicitly modeling flow in occupancy forecasting, we augment OccWorld~\cite{zheng2025occworld} to consume both the dynamic and static voxel flows provided by our unified dataset. Specifically, we introduce an additional flow encoder that processes the per-voxel flow, followed by cross-attention~\cite{vaswani2017attention} to fuse the encoded flow features with the scene tokens in the Spatial-Temporal Generative Transformer of OccWorld. We further append a flow decoder to predict next-step voxel flows, supervised via an $L_2$ loss against our ground-truth flow annotations. 

As shown in Table~\ref{table:occworld_flow} and~\ref{table:occworld_flow_waymo}, flow consistently improves forecasting performance and enhances temporal consistency for both nuScenes and Waymo datasets. In particular, we observe larger gains in the category-averaged mIoU, implying that using flow information helps the network better capture object-level motion, thereby improving predictions for dynamic classes (\textit{e.g.}, moving vehicles and pedestrians).

\subsection{Cross Data Source Training and Evaluation for Occupancy Forecasting}
\label{subsec:ood}

A key advantage of our unified dataset is the ability to train and evaluate models across multiple data sources, thereby measuring out-of-distribution (OOD) performance. We use our flow-augmented OccWorld~\cite{zheng2025occworld} to illustrate this cross-domain generalization, as it can be trained on large datasets with minimal computational overhead. Our results are shown in Table \ref{table:occworld}. The key insights are summarized below.

\noindent\textbf{Diverse Data Improves OOD Generalization in Real-World Driving.}
Models trained on a single data source (\textit{e.g.}, only nuScenes) tend to perform well on in-domain test data but exhibit weaker transfer to unseen domains (\textit{e.g.}, Waymo). In contrast, the model trained on our unified occupancy dataset (combining nuScenes and Waymo) consistently achieves higher mIoU and IoU scores over a range of prediction horizons. This implies the importance of multi-domain coverage: exposure to a broader set of scenes and motion patterns reduces the severity of domain shift and improves OOD performance. Consequently, our unified dataset marks a substantial step toward more robust occupancy forecasting in real-world driving.

\noindent\textbf{Diverse Data Improves Long Term Accuracy.}
As expected, forecast accuracy degrades with increased time horizons 
(\SI{1}{s}-\SI{3}{s}), highlighting the challenge of long-horizon occupancy prediction. Yet, this degradation is consistently less severe for the models that are trained from multi-domain data (nuScenes and Waymo), which indicates that diverse training data helps improve forecasting accuracy over time.

\begin{figure}[!]
\includegraphics[width=\linewidth]{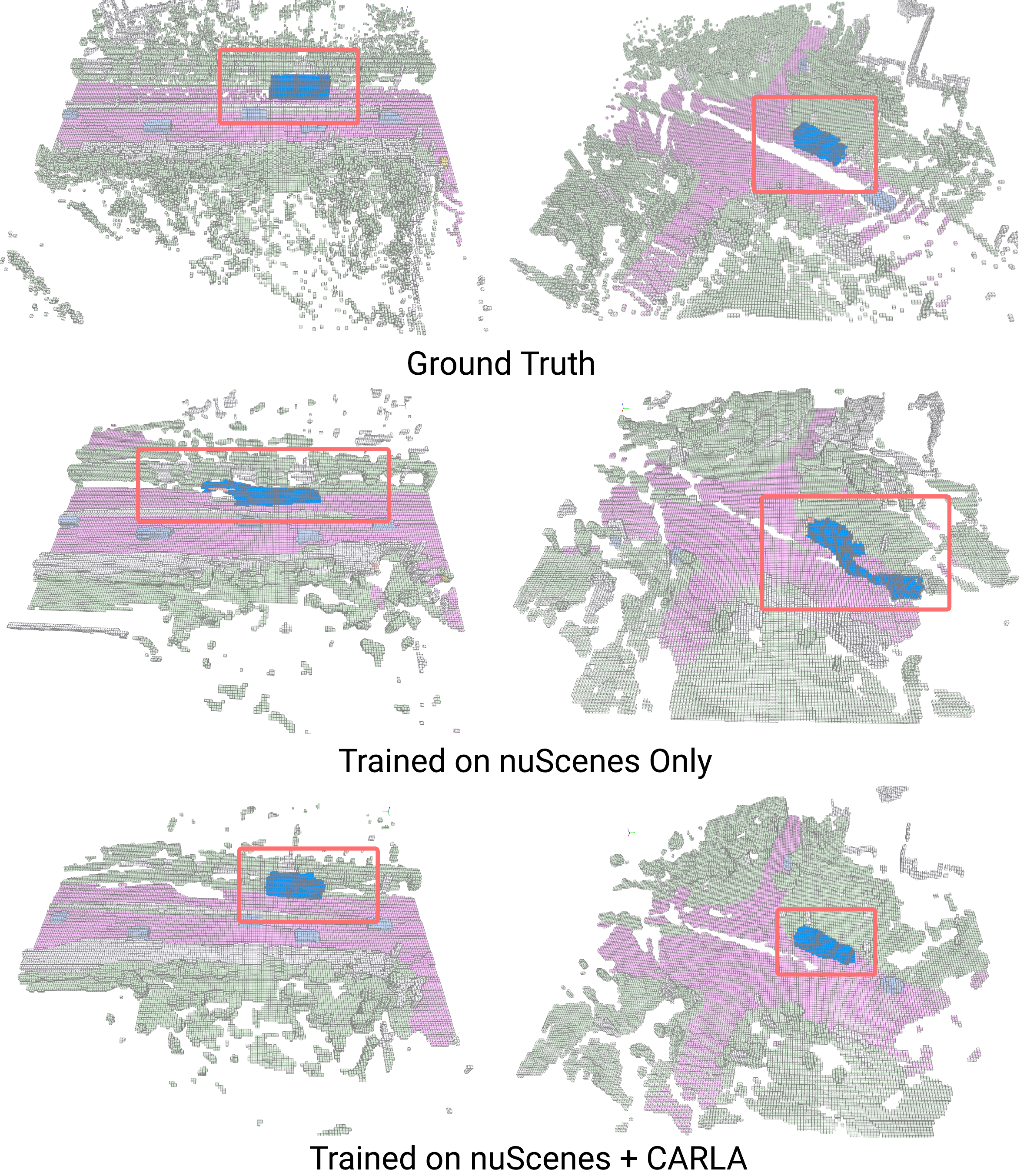}
\vspace*{-4mm}
\caption{Comparisons of the predicted car shapes between the models trained on nuScenes-only and nuScenes + CARLA data sources. In both cases, we observe a vehicle with a strange shape in the nuScenes-only forecasting. In contrast, the model trained on both data sources generates more reasonable predictions.}
\vspace{-0.cm}
\label{fig:shape}
\centering
\end{figure}

\noindent\textbf{Simulation Data Enhances Object Shape Learning.}
As is shown in Table~\ref{table:occworld}, incorporating CARLA data alongside real-world datasets increases the likelihood of accurate object predictions (noted by higher $\textbf{P}_\text{car}$), especially in scenarios where object shapes are imperfectly captured by LiDAR or pseudo-labeling in nuScenes and Waymo datasets (see Figure~\ref{fig:gt_flaw}). The simulation data, by providing ``perfect'' shapes for both static and dynamic elements, enables better learning of the geometry of objects. Figure~\ref{fig:shape} provides a qualitative illustration of this improvement.

\noindent\textbf{Sim-Real Compatibility Remains an Open Challenge.}
As is shown in Table~\ref{table:occworld}, incorporating CARLA data alongside real-world datasets increases the likelihood of accurate object predictions (noted by higher $\textbf{P}_\text{car}$), especially in scenarios where object shapes are imperfectly captured by LiDAR or pseudo-labeling in nuScenes and Waymo datasets (see Figure~\ref{fig:gt_flaw}). The simulation data, by providing ``perfect'' shapes for both static and dynamic elements, enables better learning of the geometry of objects. Figure~\ref{fig:shape} provides a qualitative illustration of this improvement. However, we also note that a naive mix of simulation data and real driving degrades temporal consistency (after adding CARLA to nuScenes, $\textbf{IoU}_{\text{bg}}$ degrades by 7.5\%). We found that one potential cause is the mismatch in speed range between CARLA and real-world datasets, since the former tends to be slower on average. We explore this perspective further in Section \ref{sim_real} in the supplementary materials.


\subsection{Occupancy Prediction}
Although our UniOcc framework primarily targets 3D occupancy forecasting from unified voxel grids, it also facilitates camera-based models, provided that domain-specific calibration and pre-processing pipelines are carefully integrated. 
We incorporate two open-source camera-based occupancy prediction approaches, Cam4DOcc~\cite{ma2024cam4docc} and CVTOcc~\cite{ye2024cvtocc}, by aligning their input requirements and output grids with our unified evaluation protocol. Notably, we evaluate each model only on the domain where its official weights were originally trained (\textit{e.g.}, Cam4DOcc on nuScenes), which enables a fair comparison of methods on a consistent voxel labeling and metric setup. Table~\ref{table:occ_pred} illustrates that CVTOcc achieves notably higher object-centric IoU and geometry-aware mIoU than Cam4DOcc due to its more flexible cost-volume fusion mechanism.

\begin{table}[!]
\vspace{0.5cm} 
\centering
\caption{Occupancy prediction performance of Cam4DOcc \cite{ma2024cam4docc} and CVTOcc \cite{ye2024cvtocc} on the nuScenes data source.}
\vspace{-0.3cm}
\label{table:occ_pred}
\resizebox{\linewidth}{!}{
\begin{tabular}{P{2.5cm}|cc|ccc}
\toprule

\textbf{Model} & $\textbf{mIoU}_{\text{geo}}$\(\uparrow\) & $\textbf{IoU}_{\text{geo}}$\(\uparrow\) & $\textbf{IoU}_{\text{bg}}$\(\uparrow\) & $\textbf{IoU}_{\text{car}}$\(\uparrow\) & $\textbf{P}_{\text{car}}$\(\uparrow\) \\

\midrule

CVTOcc \cite{ye2024cvtocc} & 31.57 & 81.20 & 48.93 & 80.60 & 74.91 \\
\midrule

Cam4DOcc \cite{ma2024cam4docc} &
13.59 & 
13.33 & 
52.46 & 56.13 & 73.28 \\
\bottomrule
\end{tabular}
}
\end{table}

\begin{table}[!]
\centering
\caption{Cooperative occupancy prediction performance of CoHFF \cite{song2024cohff} on the OpenCOOD \cite{xu2022opv2v} data source.}
\vspace{-0.3cm}
\label{table:coop_pred}
\resizebox{\linewidth}{!}{
\begin{tabular}{P{2.5cm}|cc|ccc}
\toprule

\textbf{Model} & $\textbf{mIoU}_{\text{geo}}$\(\uparrow\) & $\textbf{IoU}_{\text{geo}}$\(\uparrow\) & $\textbf{IoU}_{\text{bg}}$\(\uparrow\) & $\textbf{IoU}_{\text{car}}$\(\uparrow\) & $\textbf{P}_{\text{car}}$\(\uparrow\) \\

\midrule

CoHFF \cite{song2024cohff} & 34.16 & 50.46 & 51.90 & 87.22 & 66.19 \\
\bottomrule
\end{tabular}
}
\end{table}


\subsection{Cooperative Occupancy Prediction}

While most existing occupancy methods focus on single-ego perception, multi-vehicle collaboration offers a promising avenue for enhanced scene understanding. To highlight this, we integrate and evaluate CoHFF~\cite{song2024cohff}, a cooperative occupancy prediction approach, within our framework. By sharing sensor observations and intermediate features across multiple agents, CoHFF mitigates occlusions and extends coverage in complex driving scenarios. 
Table~\ref{table:coop_pred} reports the performance of CoHFF on the OpenCOOD~\cite{xu2022opv2v} data source, showing that multi-agent fusion yields reasonably high IoU for car instances (87.22) and background occupancy (51.90). 
These results demonstrate the potential benefits of cooperative perception and underscore our framework’s flexibility in accommodating multi-agent settings with standardized occupancy representations.

\section{Conclusion}
\label{conclusions}

We present UniOcc, a unified benchmark for occupancy forecasting and prediction in autonomous driving. By integrating diverse real-world and synthetic data sources, our approach enables cross-dataset training and evaluation on occupancy tasks ranging from single-ego to cooperative multi-vehicle settings. Beyond occupancy grid labels, we provide comprehensive occupancy flow annotations (forward and backward), voxel-based segmentation and tracking tools, and ground-truth-free evaluation metrics. We release our benchmark to foster new opportunities in the exploration of occupancy-based autonomous driving.

{
    \small
    \bibliographystyle{abbrv}
    \bibliography{main}
}

\clearpage
\setcounter{page}{1}
\maketitlesupplementary

\section{Broader Impacts}
Our unified occupancy framework, UniOcc, holds significant promise for improving safety and efficiency in autonomous driving systems. By standardizing occupancy labels and providing voxel-level flow annotations across real and simulated domains, UniOcc supports a broad spectrum of perception and forecasting tasks. This paves the way for more robust multi-modal reasoning and trajectory planning.

 Large multi-modal foundation models become increasingly influential in autonomous driving~\cite{xing2025openemma, xing2025can}. However, their training requires a large amount of data~\cite{kaplan2020scaling}. Our dataset, by unifying large-scale occupancy-image data, enables the deeper exploration of occupancy-image based foundation models~\cite{hetang2023novel,liu2024toward}. Furthermore, by enabling cross-domain evaluation and providing the toolkit for occupancy understanding, our dataset establishes a more interactive, transparent, and trustworthy~\cite{xing2024autotrust, xing2025re} framework for next-generation autonomous driving.

Our work aligns with a broader research agenda that spans generative and discriminative modeling for autonomous driving. For instance, the incorporation of occlusion-free LiDAR from aerial-ground collaborative platforms like AirV2X~\cite{gao2025airv2x} and novel view synthesis \cite{hetang2023novel} can further enhance scene understanding in the occupancy space. The ability to reason about partial observations, as explored in point-based reasoning frameworks~\cite{zhang2021point, lin2025strong, zhang2025gps, lin2023infocd, lin2023hyperbolic, kang2025lp, deng2024covis}, complements UniOcc's handling of voxel-level sparsity. Moreover, the integration of large vision-language models (VLMs) for driving scene understanding~\cite{xing2025can,xing2025re, ma2025position, pan2024plum, li2025safeflow, xing2025demystifying, zhuang2025exploring} opens new directions for using language and retrieval-augmented feedback in occupancy prediction. Relatedly, our efforts in benchmarking trustworthiness in VLMs for driving~\cite{xing2024autotrust, xing2025openemma, gao2025langcoop} can guide future work on safety-critical deployment of such models in open-world environments.

Finally, the unified occupancy modeling in UniOcc provides a strong foundation for generative world models~\cite{wang2025generative, liu2024toward} and equivariant scene reasoning~\cite{wang2023eqdrive,wang2023equivariant}, which are essential for long-horizon forecasting and generalization across cities and sensor setups. We believe that UniOcc will accelerate the development and evaluation of both classical and foundation-model-driven autonomous driving systems.

\section{Algorithm Details}

\subsection{Occupancy Flow Computation}

We discuss the details of our occupancy flow computation introduced in Section~\ref{sec:flow}.

\begin{enumerate}[leftmargin=*]
    \item \textbf{Dynamic Foreground.} At time $t$, we gather all voxels coordinates $V_{n,d}^t \in \mathbb{R}^{n \times 3}$ that belong to a labeled agent $a$. We also have the agent-to-ego transformations at times $t$ and $t+1$, namely $T_{a,t}^{e,t}$ and $T_{a,t+1}^{e,t+1}$, as well as the ego-to-ego transformation $T_{e,t}^{e,t+1} = (T_{e,t+1}^{w})^{-1} T_{e,t}^{w}$. We compute the agent’s frame transformation as:
    \begin{equation}
        T_{a,t}^{a,t+1} 
        = \bigl(T_{a,t+1}^{e,t+1}\bigr)^{-1}\;T_{e,t}^{e,t+1}\;T_{a,t}^{e,t}, 
    \end{equation}
    whose inverse gives the per-voxel motion in the $a, t$ frame:
    \begin{equation}
        M_{a,t}^{a,t+1} = \bigl(T_{a,t}^{a,t+1}\bigr)^{-1}.
    \end{equation}
    We then transform each voxel $V_{n,d}^t$ to predict its position $\widetilde{V_{n,d}^{t+1}}$ at $t+1$:
    \begin{equation}
       \widetilde{V_{n,d}^{t+1}} 
       = T_{a,t+1}^{e,t+1} \;
         T_{a,t}^{a,t+1} \;
         M_{a,t}^{a,t+1} \;
         T_{e,t}^{a,t} \;
         V_{n,d}^t.
    \end{equation}
    The flow field is then given by:
    \begin{equation}
    \label{flow_def}
       F_{n,d}^t =  \widetilde{V_{n,d}^{t+1}} - V_{n,d}^t,
    \end{equation}
    expressed in the ego coordinate frame by default, though we optionally provide an agent-centric flow variant for rotation-invariant models~\cite{zhou2022hivt,wang2023eqdrive}.

    \item \textbf{Static Background.} For each static voxel $V_{n,s}^t$, we compute its coordinates at $t+1$ using:
    \begin{equation}
       \widetilde{V_{n,s}^{t+1}} 
       = \bigl(T_{e,t+1}^{w}\bigr)^{-1} \; T_{e,t}^{w} \; V_{n,s}^t,
    \end{equation}
    and obtain $F_{n,s}^t$ as in Eq.~\eqref{flow_def}.
\end{enumerate}

For 2D flows, we begin with 2D grids and apply the same transformation steps outlined above. 

\subsection{GMM for Shape Comparison}
\label{appendix:gmm}

We discuss the details of the GMM model fitting process in this section.

Let $D_c\in \mathbb{R}^{n \times 3}$ be the $n$ dimensions for a category $c$ (for example \textit{Car}). We wish to fit a GMM with \(K\) components. Each GMM is parameterized by:
\[
\text{GMM}_c: \{\pi_k, \boldsymbol{\mu}_k, \mathbf{\Sigma}_k\}_{k=1}^K,
\]
where
\begin{itemize}
    \item \(\pi_k\) are the mixing coefficients, with \(\sum_{k=1}^K \pi_k = 1\).
    \item \(\boldsymbol{\mu}_k \in \mathbb{R}^3\) are the mean vectors.
    \item \(\mathbf{\Sigma}_k \in \mathbb{R}^{3 \times 3}\) are the covariance matrices (usually symmetric and positive-definite).
\end{itemize}

We fit the GMM parameters via the standard expectation-maximization(EM) algorithm so that each GMM component is “pulled” toward the points with high responsibility for the component. Rather than manually setting the number of components \( K \) and the covariance type, we perform a hyperparameter search using the Bayesian Information Criterion (BIC). Given a dataset of bounding box dimensions for a category, we fit multiple GMMs with varying numbers of components \( K \) and covariance types to find the optimal configuration.

Let \( \mathcal{B} \) be the set of extracted bounding box dimensions:
\[
\mathcal{B} = \{<l_i, w_i, h_i> \mid i = 1, \dots, n\},
\]
where \( l, w, h \) denote length, width, and height. To introduce robustness to the dimension data from occupancy predictions, we apply slight random perturbations corresponding to the voxel resolution:
$$
l \leftarrow l + \epsilon_l, \quad w \leftarrow w + \epsilon_w, \quad h \leftarrow h + \epsilon_h, \quad \epsilon \sim \mathcal{U}(-\frac{\epsilon}{2}, \frac{\epsilon}{2}).
$$

We then perform a grid search over:
\begin{itemize}
    \item \( K \in \{1, 2, \dots, 20\} \) (number of components)
    \item Covariance types: \textit{spherical}, \textit{tied}, \textit{diag}, \textit{full}
\end{itemize}

For each candidate model, we compute the BIC score and select the best configuration:
$$
\hat{K}, \hat{\Sigma} = \arg\min_{K, \Sigma} \text{BIC}(K, \Sigma).
$$
This ensures that the selected GMM balances model complexity and data fit. With the learned GMM, we can evaluate how well each detected object fits the learned shape distribution, enabling object classification and outlier identification. 

\begin{figure}
    \centering
    \includegraphics[width=\linewidth]{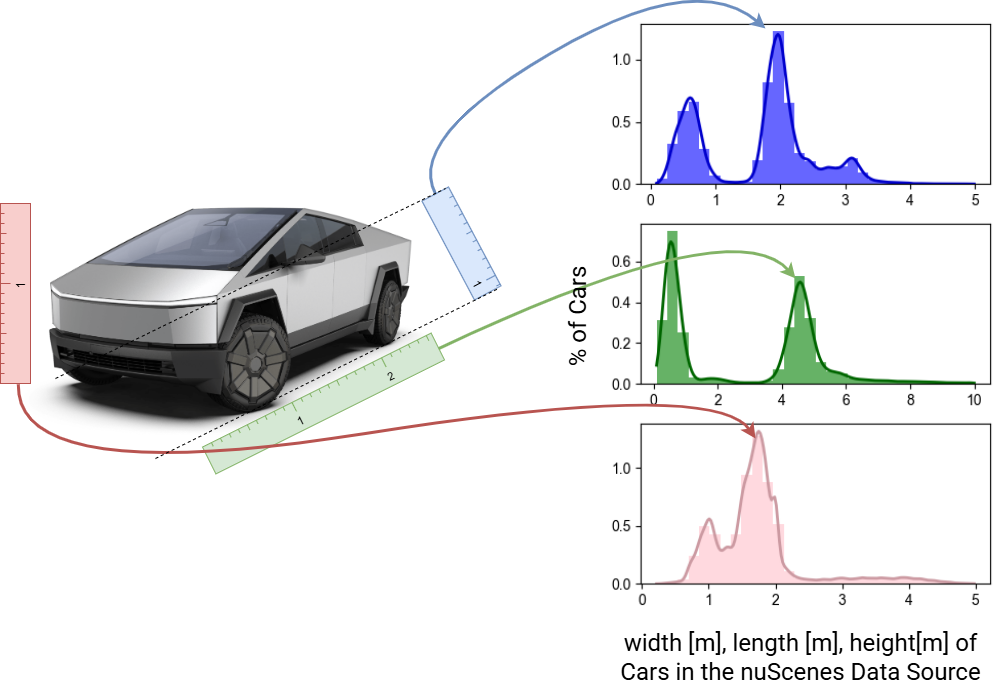}
    \caption{We compute the probability of a car by comparing its dimensions to the dimensional distribution of all cars in the data source.}
    \label{gmm_match}
\end{figure}

Once we acquire the GMM distribution, we measure the dimension of the predicted/forecasted objects and compare it with the distribution to estimate its probability of being a true object corresponding to that category. Figure~\ref{gmm_match} illustrates this process.

\subsection{Ego Localization in the Occupancy Space}
\label{localization}

\begin{figure}
    \centering
    \includegraphics[width=0.6\linewidth]{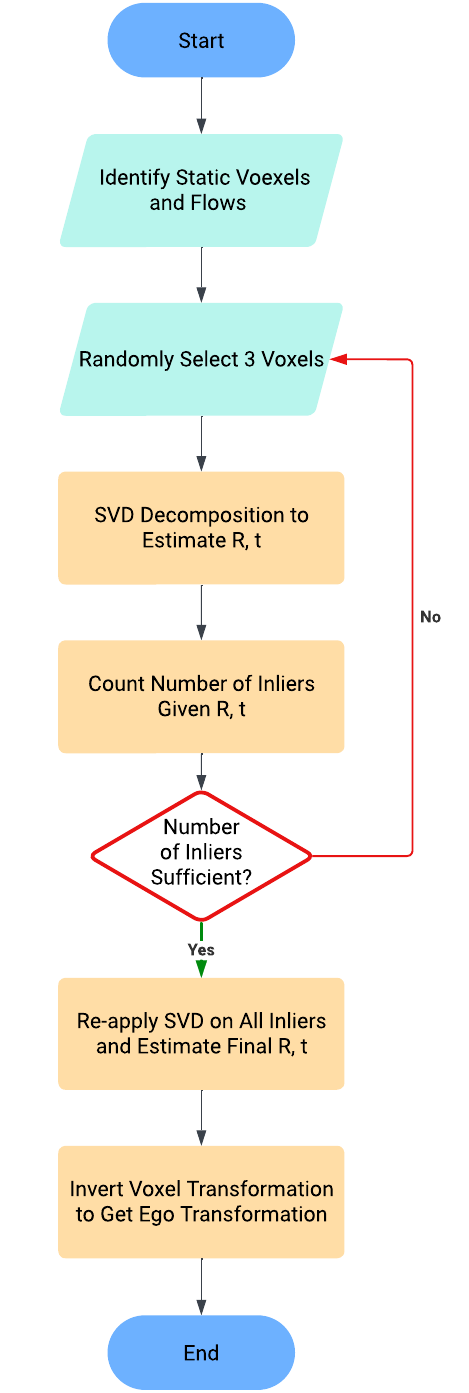}
    \caption{Occupancy space localization steps.}
    \label{localization_flowchart}
\end{figure}

By design, the results of 3D occupancy forecasting do not inherently include the future poses of the ego vehicle. However, such information is critical for downstream tasks such as object tracking, behavior prediction, and motion planning. To address this gap, we propose a method to estimate the ego motion directly from a sequence of occupancy grids and associated flow fields. Specifically, we leverage the 3D flow vectors observed on static voxels (\textit{e.g.}, road), under the assumption that their apparent motion arises solely from the ego vehicle's movement between frames. 

Given a temporal sequence of dense 3D occupancy grid $G \in \{0, \ldots, C\}^{T \times L \times W \times H}$ and a corresponding flow field $F \in \mathbb{R}^{T \times L \times W \times H \times 3}$, we isolate static scene elements and estimate the ego vehicle's $SE(3)$ transformation via robust point cloud registration. This procedure is performed independently at each frame to recover the frame-to-frame ego motion.

\subsubsection{Estimating Rigid Transform from 3D Motion}

Given that each voxel in the forecasted 3D occupancy grid is associated with a semantic class, we can isolate voxels corresponding to static scene elements (e.g., roads, buildings). Let $V_s^t$ denote the set of static voxels at time $t$, and let $F_s^t$ represent the corresponding 3D flow vectors.

According to our flow definition in Eq.~\eqref{flow_def}, the estimated position of each static voxel in the next frame can be computed as:

\begin{equation}
\widetilde{V_{s}^{t+1}} = V_s^t + F_s^t
\end{equation}

We aim to estimate a rigid-body transformation $(\mathbf{R}, \mathbf{t}) \in \mathrm{SE}(3)$ that aligns the original static voxel set with its predicted next-frame positions:

\begin{equation}
\widetilde{V_{s}^{t+1}} \approx \mathbf{R} V_s^t + \mathbf{t}
\end{equation}

Similar to the centroid extraction in Eq.~\eqref{centroid_computation}, we compute the centroids of the voxels. We then re-center them by subtracting the centroids, and we denote them as $\overline{\widetilde{V_{s}^{t+1}}}$, $\overline{V_s^t}$.

To estimate the optimal rigid transformation, we apply the Kabsch algorithm:

\begin{enumerate}[leftmargin=*]
  \item Compute cross-covariance matrix:
  \[
  \mathbf{H} = \sum_{i=1}^N \overline{\widetilde{V_{s}^{t+1}}} \overline{V_s^t}^\top
  \]
  \item Perform SVD: \( \mathbf{H} = \mathbf{U} \boldsymbol{\Sigma} \mathbf{V}^\top \)
  \item Recover rotation:
  \[
  \mathbf{R} = \mathbf{V} \mathbf{U}^\top
  \]
  \item If \( \det(\mathbf{R}) < 0 \), negate the third column of \( \mathbf{V} \) to ensure a proper rotation.
  \item Compute translation:
  \[
  \mathbf{t} = \widetilde{V_{s}^{t+1}} - \mathbf{R} V_s^t
  \]
\end{enumerate}

\subsubsection{Robust Estimation with RANSAC}

To improve robustness against noise and outliers in the flow field, we adopt a RANSAC-based procedure for estimating the ego-motion transform. At each iteration, we perform the following steps:

\begin{enumerate}[leftmargin=*]
  \item Randomly sample 3 point-motion pairs $V_{s,3}^t$, $F_{s,3}^t$.
  \item Estimate a rigid transform \( (\mathbf{R}, \mathbf{t}) \) like above.
  \item Compute residuals:
  \[
  \varepsilon_i = \left\| \mathbf{R} V_{s,3}^t + \mathbf{t} - \widetilde{V_{s, 3}^{t+1}} \right\|_2
  \]
  \item Count inliers with \( \varepsilon_i < \tau \), where \( \tau \) is a fixed threshold.
  \item Retain the transform with the largest inlier count.
\end{enumerate}

Once RANSAC converges, we re-estimate the final rigid transformation using all inliers to improve accuracy. In our experiments, we set the inlier threshold \( \tau = 0.01 \) and run for a maximum of 100 iterations, which provides a good trade-off between accuracy and efficiency.

\subsubsection{Ego Motion via Inversion}

The estimated transformation \( (\mathbf{R}, \mathbf{t}) \) maps voxel points from time \( t \) to \( t+1 \). Since the scene is assumed static, this transform is the inverse of the ego motion, which is therefore:

\[
\mathbf{R}_{\text{ego}} = \mathbf{R}^\top, \quad 
\mathbf{t}_{\text{ego}} = -\mathbf{R}^\top \mathbf{t}
\]

This yields the relative ego pose between frames \( t \) and \( t+1 \), expressed in the ego's initial frame. The steps above are illustrated in Figure~\ref{localization_flowchart}.

\subsubsection{Accumulating Camera Poses}

To obtain the ego’s trajectory, we recursively compose relative ego motion transforms:

\[
\hat{\mathbf{T}}_0 = \mathbf{I}_{4 \times 4}, \quad 
\hat{\mathbf{T}}_{t+1} = \hat{\mathbf{T}}_t \cdot \mathbf{T}_{t+1}
\]

This yields the ego pose \( \hat{\mathbf{T}}_t \in \mathrm{SE}(3) \) in the coordinate frame of the initial timestep.

\subsection{Occupancy Space Tracking, Alignment and Comparison Process}

\begin{figure}[h!]
    \centering
    \includegraphics[width=0.95\linewidth]{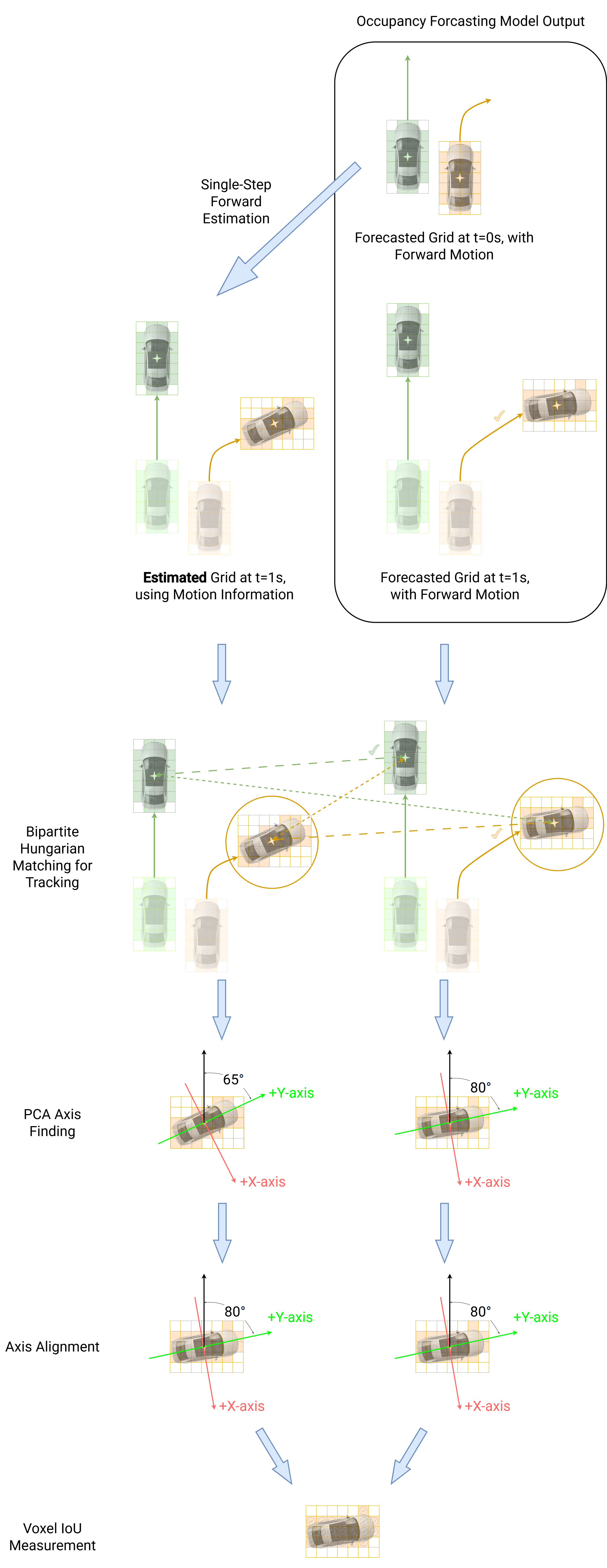}
    \caption{Flow Chart for Occupancy Space Tracking, Alignment and Comparison.}
    \label{fig:occ_track}
\end{figure}

\begin{figure}
    \centering
    \includegraphics[width=0.8\linewidth]{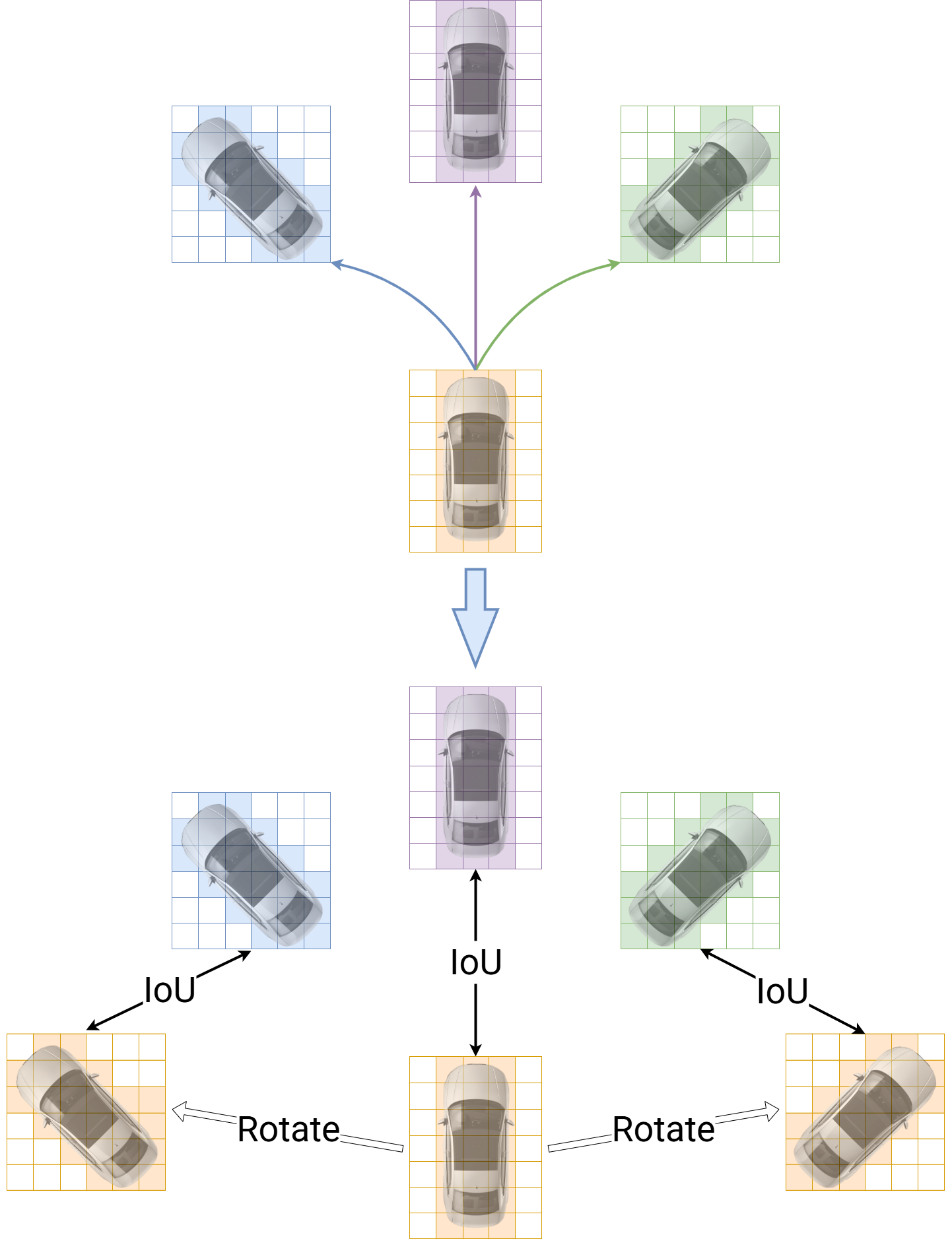}
    \caption{Example of how temporal consistency is measured in a multi-modal of three futures. \textit{I.e.}, the ego is rotated to match each of the modalities.}
    \label{fig:multi-modal_future}
\end{figure}

In Section~\ref{sec:object_tracking}, we discuss the process for occupancy tracking, alignment, and shape comparison. In Figure~\ref{fig:occ_track}, we further illustrate it with an example. We begin with two consecutive model-forecasted occupancy grids (e.g., from OccWorld~\cite{zheng2025occworld}) at \( t = 0\,\text{s} \) and \( t = 1\,\text{s} \), each containing voxel-wise semantic labels and flow vectors. In this example, we choose \textit{car} as an example, so we isolate the voxels with semantic label \textit{car}. Using the occupancy grid at \( t = 0\,\text{s} \), we perform a single-step forward estimation by adding the forward flow vectors to the occupied voxels to obtain their predicted coordinates in the next frame.

Next, we compute the centroids of all segmented objects in both the estimated and forecasted grids. We apply bipartite Hungarian matching to associate corresponding object instances across frames, completing the tracking step. To evaluate shape consistency, we apply the alignment procedure described in Algorithm~\ref{algo:pca_consistent}, which aligns the voxel sets by rotating the estimated objects to match the forecasted ones using principal axis alignment. Once aligned, we measure their voxel-wise similarity via the intersection-over-union (\textbf{IoU}) metric.

Importantly, this method of computing temporal consistency is modality-agnostic. For instance, in Figure~\ref{fig:multi-modal_future}, we illustrate the case of multi-modal forecasting with three possible futures. Temporal consistency can be independently computed for each future branch by aligning the prior frame’s occupied voxels to the corresponding forecasted voxels before evaluating their overlap.

It is also important to note that the tracking procedure applies not just to the ego, but also to the other agents, allowing us to turn the occupancy forecasting problem into a motion prediction problem.

\subsection{Voxel Alignment with PCA}

In Algorithm \ref{algo:pca_consistent}, we include the pseudo code to align voxels through the temporal dimension by standardizing their rotations, which we introduced in Section \ref{sec:pca}.

\begin{algorithm}
    \SetAlgoLined
    \KwIn{\(V^t\): A set of 3D points at timestep \( t \), \(\text{pca}^{t-1}\) (if available)}
    \KwOut{\(\hat{V}^t\): Transformed coordinates with consistent orientation}
    
    \BlankLine
    \emph{Step 1: Fit the PCA to the current timestep.}\\
    \(\text{pca}^t.\mathrm{fit}(V^t)\) \\    
    $\text{pca}_{\mathrm{ref}} = \text{pca}^t$
    
    \BlankLine
    \emph{Step 2: Transform coordinates into the principal component space.}\\
    \(\hat{V}^t \leftarrow \text{pca}^t.\mathrm{transform}(\bar{V}^t)\)
    
    \BlankLine
    \For{\(t \in \{t_2, \dots, t_n\}\)}{
        \emph{Step 3: Ensure consistent handedness using PCA from previous timestep.}\\

        \emph{(a) Compute diagonal sign vector.}\\
        \(k^t = \text{pca}^t.\mathrm{components} \) \\
        \( k_{ref}^{t-1} = \text{pca}^{t-1}_{\mathrm{ref}}.\mathrm{components}^T  \)\\
        \(\boldsymbol{\sigma} \leftarrow \mathrm{sign} \Big(\mathrm{diag} \big( k^t \cdot  k_{ref}^{t-1} \big) \Big)\)
        
        \BlankLine
        \emph{(b) Apply sign correction to maintain consistency.}\\
        \(\hat{V}^t \leftarrow \hat{V}^t \odot \boldsymbol{\sigma}\) 
        \tcp*[r]{Hadamard product ensures consistent PCA orientation}
    }

    \Return \(\hat{V}^t\)
    
    \caption{PCA-Based Coordinate Transformation with Consistent Handedness}
    \label{algo:pca_consistent}
\end{algorithm}

\section{Sim-Real Compatibility}
\label{sim_real}

\begin{table*}[h!]
\centering
\caption{Occupancy forecasting performance of OccWorld \cite{zheng2025occworld} after mixing into the training data with either full-speed-range (Unbiased) CARLA scenes or highway-only CARLA scenes.}
\label{table:carla_speed_exp}
\resizebox{\textwidth}{!}{
\begin{tabular}{c|c|cccc|cccc|ccc}
\toprule
\textbf{Train Sources} & \textbf{Test Source} & 
\multicolumn{4}{c|}{$\textbf{mIoU}_{\text{geo}}$\(\uparrow\)} &
\multicolumn{4}{c|}{$\textbf{IoU}_{\text{geo}}$\(\uparrow\)} &
\multicolumn{1}{c}{$\textbf{IoU}_{\text{bg}}$\(\uparrow\)} &
\multicolumn{1}{c}{$\textbf{IoU}_{\text{car}}$\(\uparrow\)} &
\multicolumn{1}{c}{$\textbf{P}_{\text{car}}$\(\uparrow\)} \\
& & \SI{0}{s} & \SI{1}{s} & \SI{2}{s} & \SI{3}{s} &
\SI{0}{s} & \SI{1}{s} & \SI{2}{s} & \SI{3}{s} &
\multicolumn{3}{c}{\text{Average over } \SIrange{0}{3}{s}}  \\
\midrule
 
nuScenes + Unbiased CARLA & nuScenes  &
71.47 & 31.70 & 22.69 & 18.11 & 
62.94 & 35.83 & 28.29 & 21.03 & 
55.07 & 77.87 & 82.97 \\

nuScenes + Highway CARLA & nuScenes &
72.63 & 31.89 & 22.21 & 21.85 & 
63.70 & 36.24 & 28.68 & 21.15 & 
61.40 & 85.59 & 82.89 \\

\midrule

Waymo + Highway CARLA & Highway CARLA  &
84.02 & 54.12 & 53.07 & 52.40 & 
74.32 & 28.11 & 24.15 & 23.05 & 
87.43 & 92.53 & 81.62 \\

nuScenes + Waymo + Highway CARLA & Highway CARLA &
87.50 & 67.15 & 61.86 & 59.75 & 
74.85 & 28.57 & 25.20 & 23.61 & 
88.14 & 98.41 & 82.54 \\

\bottomrule
\end{tabular}
}
\end{table*}

The simulation data generated from CARLA differs from real-world driving data in several important aspects, including ego and agent speeds, traffic density, and vehicle geometry. Among these factors, we found that differences in agent speed between simulation and real-world datasets have the most significant impact on the temporal consistency of the trained models. Specifically, scenes in our CARLA dataset tend to feature narrower streets and slower-moving agents compared to those in nuScenes or Waymo. A comparative analysis of speed distributions is shown in Figure~\ref{speed_dist}.

To quantify the impact of this discrepancy, we conducted additional experiments using a subset of CARLA scenes focused on highway driving, generated with the \textit{Town06} map, which supports speeds up to \SI{55}{mph} (\SI{24.5}{m/s}). As shown in Table~\ref{table:carla_speed_exp}, incorporating these high-speed scenes into training yields notable improvements: +11\% in background consistency (\(\text{IoU}_\text{bg}\)) and +9\% in object-level consistency (\(\text{IoU}_\text{car}\)). These results suggest that higher-speed simulated environments lead to more realistic motion dynamics and improved temporal coherence when used for joint training.

In our future work, we plan to further explore the role of sim-to-real compatibility and investigate principled ways to adapt or calibrate simulation data for improved generalization to real-world domains.

\begin{table*}
    \centering
    \caption{Distribution of ego vehicle speeds in CARLA and nuScenes datasets. The standard CARLA scenes exhibit predominantly low-speed driving behavior, whereas nuScenes features a broader speed distribution. Our highway-focused CARLA subset (generated using the \textit{Town06} map) better matches the real-world speed range observed in nuScenes, improving sim-real compatibility for training.}
    \resizebox{\textwidth}{!}{
    \begin{tabular}{c|c|c}
        \includegraphics[]{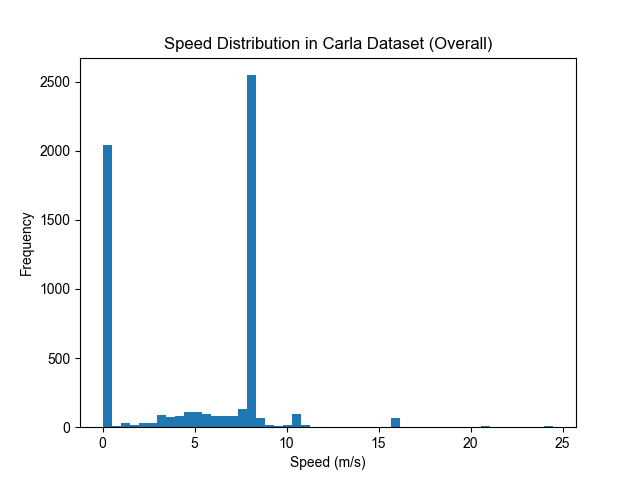} & \includegraphics[]{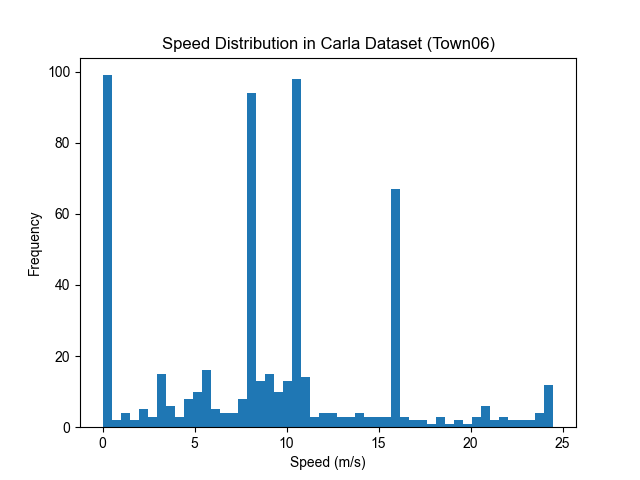} & \includegraphics[]{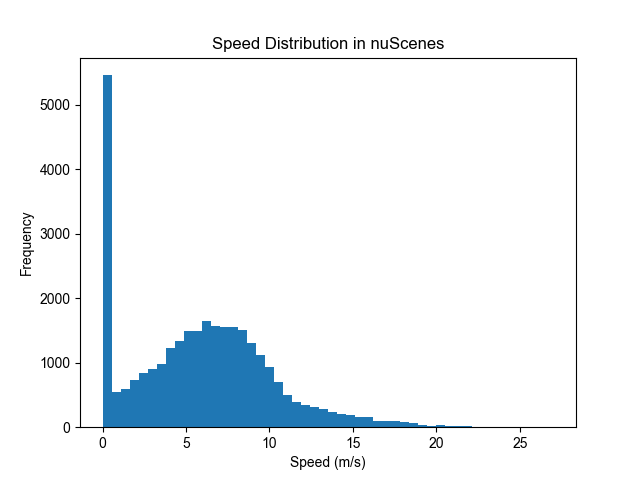} \\
    \end{tabular}        
    }
    \label{speed_dist}
\end{table*}

\begin{table*}
\centering
\caption{Occupancy forecasting performance of OccWorld \cite{zheng2025occworld} on Waymo datasets with different types of flow.}
\vspace{-0.3cm}
\label{table:occworld_flow_waymo}
\resizebox{\textwidth}{!}{
\begin{tabular}{P{44mm}|P{17mm}|cccc|cccc|ccc}
\toprule
\textbf{Train and Test Source} & \textbf{Flow Type} & 
\multicolumn{4}{c|}{$\textbf{mIoU}_{\text{geo}}$\(\uparrow\)} &
\multicolumn{4}{c|}{$\textbf{IoU}_{\text{geo}}$\(\uparrow\)} &
\multicolumn{1}{c}{$\textbf{IoU}_{\text{bg}}$\(\uparrow\)} &
\multicolumn{1}{c}{$\textbf{IoU}_{\text{car}}$\(\uparrow\)} &
\multicolumn{1}{c}{$\textbf{P}_{\text{car}}$\(\uparrow\)} \\

& & \SI{0}{s} & \SI{1}{s} & \SI{2}{s} & \SI{3}{s} &
\SI{0}{s} & \SI{1}{s} & \SI{2}{s} & \SI{3}{s} &
\multicolumn{3}{c}{\text{Average over } \SIrange{0}{3}{s}}  \\
\midrule
Waymo & None &
68.24 & 30.40 & 24.03 & 21.79 &
70.89 & 34.41 & 28.85 & 26.33 &
56.06 & 88.10 & 83.55 \\

Waymo & Object &
67.66 & 30.65 & 24.41 & 21.93 &
71.26 & 34.03 & 29.18 & 26.60 &
55.56 & 88.11 & 84.30 \\

Waymo & Voxel &
\textbf{71.35} & \textbf{32.04} & \textbf{25.77} & \textbf{23.76} &
\textbf{72.69} & \textbf{36.04} & \textbf{30.48} & \textbf{27.96} &
\textbf{58.26} &\textbf{89.30} & \textbf{86.68} \\

\bottomrule
\end{tabular}
}
\end{table*}

\section{Unified Quality Score}

Even though our proposed metrics do not depend on existing pseudo‑labels, we retain the current widely used labels/metrics to keep comparability with prior works, while our novel metrics provide further insights into shape realism and consistency of voxels. 

To facilitate direct comparison across methods and evaluation settings, we introduce a composite metric, \textbf{UniOcc Score}, which aggregates multiple specialized metrics via a weighted average:

\[
\begin{aligned}
\text{UniOccScore} =\;& 
\lambda_{\text{Recon}} \cdot \text{IoU}^{t=0\text{s}}_{\text{geo}} \\
&+ \lambda_{\text{Forecast\_1s}} \cdot \text{IoU}^{t=1\text{s}}_{\text{geo}} \\
&+ \lambda_{\text{Forecast\_2s}} \cdot \text{IoU}^{t=2\text{s}}_{\text{geo}} \\
&+ \lambda_{\text{Forecast\_3s}} \cdot \text{IoU}^{t=3\text{s}}_{\text{geo}} \\
&+ \lambda_{\text{Temp\_bg}} \cdot \text{IoU}_{\text{bg}} \\
&+ \lambda_{\text{Temp\_car}} \cdot \text{IoU}_{\text{car}} \\
&+ \lambda_{\text{Probs\_car}} \cdot \text{P}_{\text{car}}
\end{aligned}
\].

Here, each \( \lambda \) weight controls the contribution of a particular aspect of performance: reconstruction quality, forecasting accuracy at various horizons, temporal consistency for background and dynamic objects, and the realism probability for predicted cars.

While the weights may be adjusted depending on the downstream application (\textit{e.g.}, emphasizing forecasting metrics in long-horizon prediction tasks), we provide a reference configuration that balances reconstruction and forecasting:  
\begin{center}
\begin{tabular}{ll}
\toprule
Weight Component & Value \\
\midrule
$\lambda_{\text{Recon}}$          & 0.20 \\
$\lambda_{\text{Forecast\_1s}}$   & 0.15 \\
$\lambda_{\text{Forecast\_2s}}$   & 0.10 \\
$\lambda_{\text{Forecast\_3s}}$   & 0.05 \\
$\lambda_{\text{Temp\_bg}}$       & 0.30 \\
$\lambda_{\text{Temp\_car}}$      & 0.20 \\
$\lambda_{\text{Probs\_car}}$     & 0.10 \\
\bottomrule
\end{tabular}
\end{center}

The resulting UniOcc scores from our experiments are reported in Table~\ref{example_uniocc_score}.

\begin{table}[h]
\centering
\caption{Reference UniOcc Score.}
\label{example_uniocc_score}
\resizebox{\linewidth}{!}{
\begin{tabular}{c|c|c}
\toprule
\textbf{Train Sources} & \textbf{Test Source} &  \textbf{UniOcc Score} \\
\midrule

nuScenes & nuScenes & 63.99 \\

Waymo & nuScenes & 57.26\\

CARLA & nuScenes & 40.10 \\

nuScenes & Waymo & 65.36\\

Waymo & Waymo & 68.40\\

CARLA & Waymo & 41.70 \\

nuScenes & CARLA & 71.48\\

Waymo & CARLA & 71.05\\

CARLA & CARLA & 46.39\\

\midrule

nuScenes + Waymo & nuScenes & 65.53\\

nuScenes + Waymo & Waymo & 68.86\\

\midrule

nuScenes + CARLA & nuScenes & 62.24\\

Waymo + CARLA & nuScenes & 58.61\\

nuScenes + CARLA & Waymo &66.13 \\

Waymo + CARLA & Waymo & 68.45\\

nuScenes + CARLA & CARLA & 73.53\\

Waymo + CARLA & CARLA &71.00 \\

\midrule

nuScenes + Waymo + CARLA & nuScenes &63.64 \\

nuScenes + Waymo + CARLA & Waymo & 70.11\\

nuScenes + Waymo + CARLA & CARLA &70.58\\

\bottomrule
\end{tabular}
}
\end{table}

\section{Additional Experiments}

In Table~\ref{table:occworld_flow_waymo}, we show the results of using voxel-level flow on OccWorld~\cite{zheng2025occworld} on the Waymo dataset. The performance gain is consistent with the nuScenes dataset in Table~\ref{table:occworld_flow}.

\section{Additional Dataset Details}

In Table \ref{table:label_comparison}, we include a list of the different labels used in our different data sources.

\begin{table}[]
\caption{Label correspondence across four different sources (nuScenes, Waymo, CARLA, and UniOcc (Ours)). Empty cells indicate no direct counterpart.}
\label{table:label_comparison}
\centering
\resizebox{\linewidth}{!}{
\begin{tabular}{c|l|l|l|l}
\hline
\textbf{ID} & \textbf{nuScenes} & \textbf{Waymo} & \textbf{Carla} & \textbf{UniOcc (Ours)} \\
\hline
0  & general\_object          & general\_object     & free                & general\_object    \\
1  & barrier                  & vehicle             & buildings           & vehicle            \\
2  & bicycle                  & pedestrian          & fences              & bicycle            \\
3  & bus                      & sign                & other               & motorcycle         \\
4  & car                      & cyclist             & pedestrians         & pedestrian         \\
5  & construction\_vehicle    & traffic\_light      & poles               & traffic\_cone      \\
6  & motorcycle               & pole                & roadlines           & vegetation         \\
7  & pedestrian               & construction\_cone  & roads               & road               \\
8  & traffic\_cone            & bicycle             & sidewalks           & walkable/terrain   \\
9  & trailer                  & motorcycle          & vegetation          & building           \\
10 & truck                    & building            & vehicles            & free               \\
11 & drivable\_surface        & vegetation          & walls               & -                  \\
12 & other\_flat              & tree\_trunk         & trafficSigns        & -                  \\
13 & sidewalk                 & road                & sky                 & -                  \\
14 & terrain                  & walkable            & ground              & -                  \\
15 & manmade                  & -                   & -                   & -                  \\
16 & vegetation               & -                   & -                   & -                  \\
17 & free                     & -                   & -                   & -                  \\
23 & -                        & free                & -                   & -                  \\
\hline
\end{tabular}
}
\end{table}

\end{document}